\documentclass[breakmath]{config}

\title{Learning in latent spaces improves the predictive accuracy of deep neural operators}
\shorttitle{Your short title goes here} 

\author[1]{Katiana Kontolati
	\orcid{0000-0003-2027-9638}
}
\author[2]{Somdatta Goswami
	\orcid{0000-0002-8255-9080}
}
\author[2,3]{George Em Karniadakis
	\orcid{0000-0002-9713-7120}
}
\author[1]{Michael D Shields
	\orcid{0000-0003-1370-6785}
	\thanks{Corresponding author. Email: \href{mailto:michael.shields@jhu.edu}{michael.shields@jhu.edu}}
}
\affil[1]{Department of Civil and Systems Engineering, Johns Hopkins University}
\affil[2]{Division of Applied Mathematics, Brown University}
\affil[3]{School of Engineering, Brown University}

\begin{document}


\makeseistitle{
	\begin{summary}{Abstract}
	Operator regression provides a powerful means of constructing discretization-invariant emulators for partial-differential equations (PDEs) describing physical systems. Neural operators specifically employ deep neural networks to approximate mappings between infinite-dimensional Banach spaces. As data-driven models, neural operators require the generation of labeled observations, which in cases of complex high-fidelity models result in high-dimensional datasets containing redundant and noisy features, which can hinder gradient-based optimization. Mapping these high-dimensional datasets to a low-dimensional latent space of salient features can make it easier to work with the data and also enhance learning. In this work, we investigate the latent deep operator network (L-DeepONet), an extension of standard DeepONet, which leverages latent representations of high-dimensional PDE input and output functions identified with suitable autoencoders. We illustrate that L-DeepONet outperforms the standard approach in terms of both accuracy and computational efficiency across diverse time-dependent PDEs, \textit{e.g.,} modeling the growth of fracture in brittle materials, convective fluid flows, and large-scale atmospheric flows exhibiting multiscale dynamical features. 
    \end{summary}
	}
	
\section{Introduction}

Achieving universal function approximation is one of the most important tasks in the rapidly growing field of machine learning (ML). To this end, deep neural networks (DNNs) have been actively developed, enhanced and used for a plethora of versatile applications in science and engineering including image processing, natural language processing (NLP), recommendation systems, and design optimization [\cite{guo2016deep,pak2017review,brown2020language,otter2020survey,khan2021deep,kollmann2020deep}]. In the emerging field of scientific machine learning (SciML), DNNs are a ubiquitous tool for analyzing, solving, and optimizing complex physical systems modeled with partial differential equations (PDEs) across a range of scenarios, including different initial and boundary conditions (ICs, BCs), model parameters and geometric domains. Such models are trained from a finite dataset of labeled observations generated from a (generally expensive) traditional numerical solver (e.g., finite difference method (FD), finite elements (FEM), computational fluid dynamics (CFD), and once trained they allow for accurate predictions with real-time inference [\cite{berg2019data,chen2019symplectic,raissi2019physics,abdar2021review}]. 

DNNs are conventionally used to learn functions by approximating mappings between finite dimensional vector spaces. Operator regression, a more recently proposed ML paradigm, focuses on learning operators by approximating mappings between abstract infinite-dimensional Banach spaces. Neural operators specifically, first introduced in 2019 with the deep operator network (DeepONet) [\cite{lu2021learning}], employ DNNs to learn PDE operators and provide a discretization-invariant emulator, which allows for fast inference and high generalization accuracy. Motivated by the universal approximation theorem for operators proposed by Chen \& Chen [\cite{chen1995universal}], 
DeepONet encapsulates and extends the theorem for deep neural networks~\cite{lu2021learning}].
The architecture of DeepONet features a DNN, which encodes the input functions at fixed sensor points (branch net) and another DNN, which encodes the information related to the spatio-temporal coordinates of the output function (trunk net). Since its first appearance, standard DeepONet has been employed to tackle challenging problems involving complex high-dimensional dynamical systems [\cite{di2021deeponet,kontolati2022influence,goswami2022physics,oommen2022learning,cao2023deep}]. 
In addition, extensions of DeepONet have been recently proposed in the context of multi-fidelity learning [\cite{de2022bi,lu2022multifidelity,howard2022multifidelity}], integration of multiple-input continuous operators [\cite{jin2022mionet,goswami2022neural}], hybrid transferable numerical solvers [\cite{zhang2022hybrid}], transfer learning [\cite{goswami2022deep}], and physics-informed learning to satisfy the underlying PDE [\cite{wang2021learning,goswami2022physics2}].

Another class of neural operators is the integral operators, first instantiated with the graph kernel networks (GKN) introduced by \cite{li2020neural}. In GKNs, the solution operator is expressed as an integral operator of Green's function which is modeled with a neural net and consists of a lifting layer, iterative kernel integration layers, and a projection layer. GKNs were found to be unstable for multiple layers and a new graph neural operator was developed in \cite{you2022nonlocal} based on a discrete non-local diffusion-reaction equation. Furthermore, to alleviate the inefficiency and cost of evaluating integral operators, the Fourier neural operator (FNO) [\cite{li2020fourier}] was proposed, in which the integral kernel is parameterized directly in the Fourier space. The input to the network, like in GKNs, is elevated to a higher dimension, then passed through numerous Fourier layers before being projected back to the original dimension. Each Fourier layer involves a forward fast Fourier transform (FFT), followed by a linear transformation of the low-Fourier modes and then an inverse FFT. Finally, the output is added to a weight matrix, and the sum is passed through an activation function to introduce nonlinearity. Different variants of FNO have been proposed, such as the FNO-2D which performs $2$D Fourier convolutions and uses a recurrent structure to propagate the PDE solution in time, and the FNO-3D, which performs $3$D Fourier convolutions through space and time. Compared to DeepONet, FNO employs evaluations restricted to an equispaced mesh to discretize both the input and output spaces, where the mesh and the domain must be the same. The interested reader is referred to \cite{lu2022comprehensive} for a comprehensive comparison between DeepONet and FNO across a range of complex applications. Recent advancements in neural operator research have yielded promising results for addressing the bottleneck of FNO. Two such integral operators are the Wavelet Neural Operator (WNO) \cite{tripura2023wavelet} and the Laplace Neural Operator (LNO) \cite{cao2023lno}, which have been proposed as alternative solutions for capturing the spatial behavior of a signal and accurately approximating transient responses, respectively.

\begin{figure}[ht!]
\begin{center}
\includegraphics[width=0.9\textwidth]{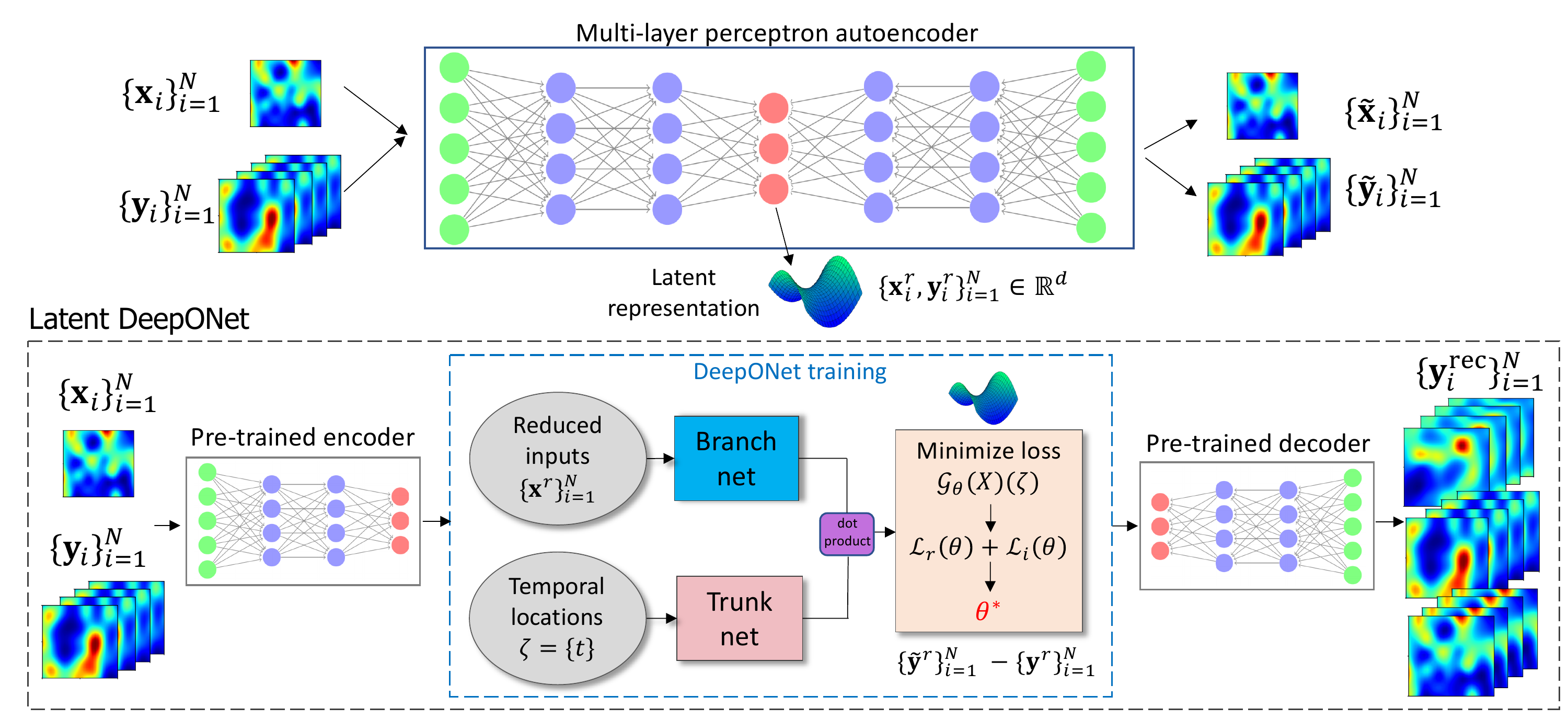}
\caption{Latent DeepONet (L-DeepONet) framework for learning deep neural operators on latent spaces. In the first step, a multi-layer autoencoder is trained using a combined dataset of the high-dimensional input and output realizations of a PDE model, $\{\mathbf{x}_i, \mathbf{y}_i\}_{i=1}^{N}$, respectively. The trained encoder projects the data onto a latent space $\mathbb{R}^d$ and the dataset on the latent space, $\{\mathbf{x}^r_i, \mathbf{y}^r_i\}_{i=1}^{N}$ is then used to train a DeepONet model and learn the operator $\mathcal{G}_{\theta}$, where $\theta$ denotes the trainable parameters of the network. Finally, to evaluate the performance of the model on the original PDE outputs and perform inference, the pre-trained decoder is employed to map predicted samples back to physically-interpretable space.}
\label{fig:L-DeepONet}
\end{center}
\captionsetup{justification=centering}
\end{figure}

Despite the impressive capabilities of the aforementioned methods to learn mesh-invariant surrogates for complex PDEs, these models are primarily used in a data-driven manner, and thus a representative and sufficient labeled dataset needs to be acquired \textit{a-priori}. Often, complex physical systems require high-fidelity simulations defined on fine spatial and temporal grids, which results in very high-dimensional datasets. Furthermore, the high (and often prohibitive) expense of traditional numerical simulators e.g., FEM allows for the generation of only a few hundred (and possibly even fewer) observations. The combination of few and very high-dimensional observations can result in sparse datasets that often do not represent adequately the input/output distribution space. In addition, raw high-dimensional physics-based data often consists of redundant features that can (often significantly) delay and hinder network optimization. Physical constraints cause the data to live on lower-dimensional latent spaces (manifolds) that can be identified with suitable linear or nonlinear dimension reduction (DR) techniques. Previous studies have shown how latent representations can be leveraged to enable surrogate modeling and uncertainty quantification (UQ) by addressing the `curse of dimensionality' in high-dimensional PDEs with traditional approaches such as Gaussian processes (GPs) and polynomial chaos expansion (PCE) [\cite{lataniotis2020extending,nikolopoulos2022non,giovanis2020data,kontolati2022manifold,kontolati2022survey}]. Although neural network-based models can naturally handle high-dimensional input and output datasets, it is not clear how their predictive accuracy, generalizability, and robustness to noise are affected when these models are trained with suitable latent representations of the high-dimensional data.

In this work, we aim to investigate the aforementioned open questions by exploring the training of DeepONet on latent spaces for high-dimensional time-dependent PDEs of varying degrees of complexity. The idea of training neural operators on latent spaces using DeepONet and autoencoders (AE) was originally proposed in \cite{oommen2022learning}. In this work, the growth of a two-phase microstructure for particle vapor deposition was modeled using the Cahn-Hilliard equation. In another recent work [\cite{zhang2022multiauto}], the authors explored neural operators in conjunction with AE to tackle high-dimensional stochastic problems. But the general questions of the predictive accuracy and generalizability of DeepONets trained on latent spaces remain and require systematic investigation with comparisons to conventional neural operators. 

The training of neural operators on latent spaces consists of a two-step approach: first, training a suitable AE model to identify a latent representation for the high-dimensional PDE inputs and outputs, and second, training a DeepONet model and employing the pre-trained AE decoder to project samples back to the physically interpretable high-dimensional space (see Figure \ref{fig:L-DeepONet}). The L-DeepONet framework has two advantages: first, the accuracy of DeepONet is improved, and second, the L-DeepONet training is accelerated due to the low dimensionality of the data in the latent space. Combined with the pre-trained AE model, L-DeepONet can perform accurate predictions with real-time inference and learn the solution operator of complex time-dependent PDEs in low-dimensional space. The contributions of this work can be summarized as follows:
\begin{itemize}
    \item We investigate the performance of L-DeepONet, an extension of standard DeepONet, for high-dimensional time-dependent PDEs that leverages latent representations of input and output functions identified by suitable autoencoders (see Figure \ref{fig:L-DeepONet}).
    \vspace{-5pt}
    \item We perform direct comparisons with vanilla DeepONet for complex physical systems, including brittle fracture of materials, and complex convective and atmospheric flows, and demonstrate that L-DeepONet consistently outperforms the standard approach in terms of accuracy and computational time.
    \vspace{-5pt}
    \item We perform direct comparisons with another neural operator model, the Fourier neural operator (FNO), and two of its variants, i.e., FNO-2D and FNO-3D, and identify advantages and limitations for a diverse set of applications.
\end{itemize}

\section{Results}
\label{sec:results}

To demonstrate the advantages and efficiency of L-DeepONet, we learn the operator for three diverse PDE models of increasing complexity and dimensionality. First, we consider a PDE that describes the growth of fracture in brittle materials which are widely used in various industries including construction and manufacturing. Predicting with accuracy the growth of fractures in these materials is important for preventing failures, improve safety, reliability and cost-effectiveness in a wide range of applications. Second, we consider a PDE describing convective fluid flow, a common phenomenon in many natural and industrial processes. Understanding how these flows evolve may allow engineers to better design systems such as heat exchangers or cooling systems to enhance efficiency and reduce energy consumption. Finally, we consider a PDE describing large-scale atmospheric flows which can be used to predict patterns that occur in weather systems. Such flows play a crucial role in the Earth's climate system influencing precipitation, temperature which in turn may have a significant impact in water resources, agricultural productivity and energy production. Developing an accurate surrogate to predict with detail such complex atmospheric patterns may allow us to better adapt to changes in the climate system and develop effective strategies to mitigate the impacts of climate change. For all PDEs, the input functions for the operator represent initial conditions modeled as Gaussian or non-Gaussian random fields. We perform direct comparisons of L-DeepONet with the standard DeepONet model trained on the full dimensional data and with FNO. More details about the models and the corresponding data generation process are provided in the Supplementary Materials to assist the readers in readily reproducing the results presented below. 

\begin{figure}[ht!]
\begin{center}
\includegraphics[width=0.7\textwidth]{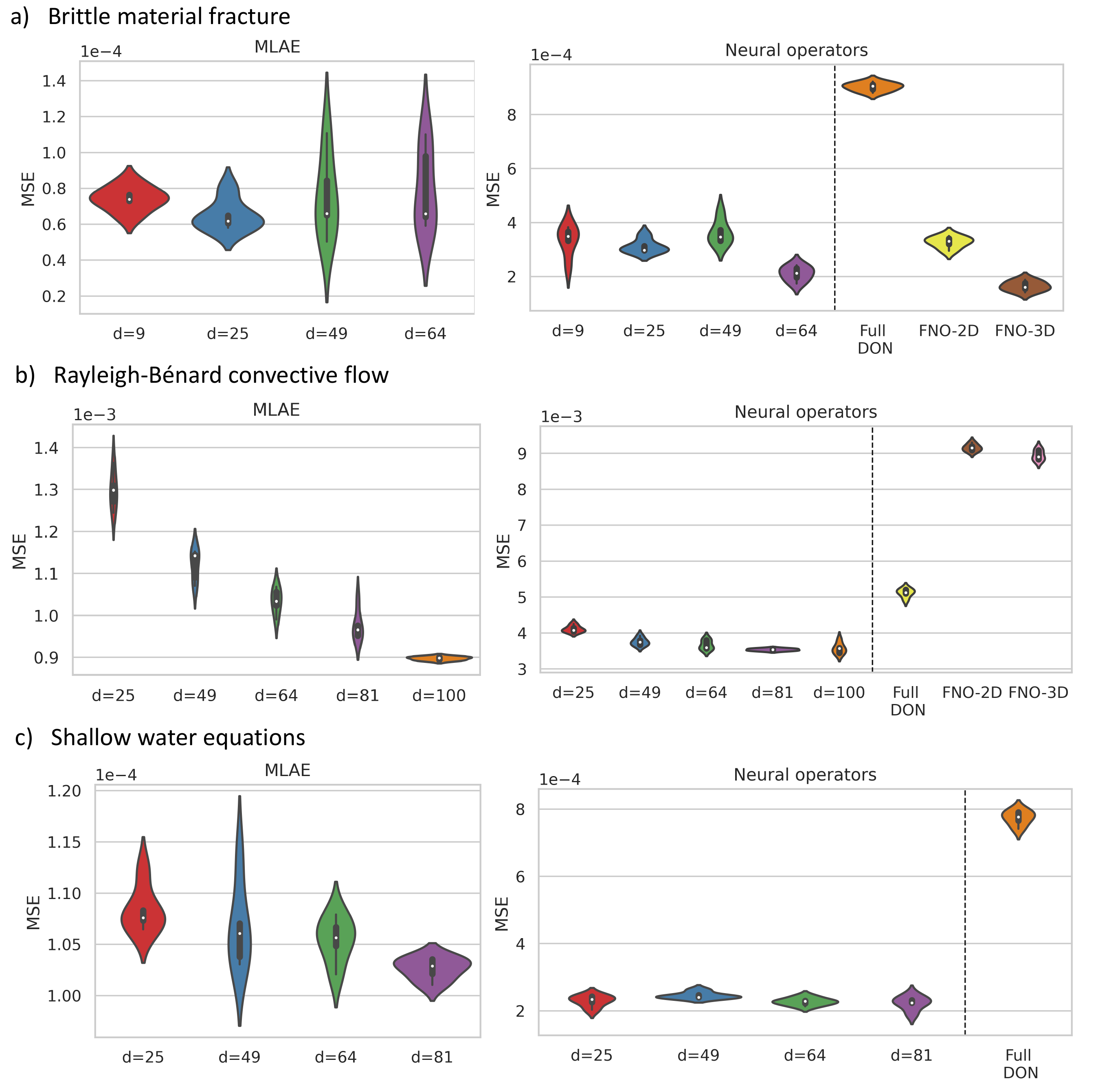}
\caption{Left: Results for all applications of the multi-layer autoencoders (MLAE) for different values of the latent dimensionality. Right: Results for all applications of the neural operators for all studied models. Violin plots represent $5$ independent training of the models using different random seed numbers.}
\label{fig:L-DeepONet-results}
\end{center}
\captionsetup{justification=centering}
\end{figure}

\bigbreak
\noindent
\textbf{Brittle fracture in a plate loaded in shear}

\noindent Fracture is one of the most commonly encountered failure modes in engineering materials and structures. Defects, once initialized, can lead to catastrophic failure without warning. Therefore, from a safety point of view, prediction of the initiation and propagation of cracks is of utmost importance. In the phase field fracture modeling approach, the effects associated with crack formation, such as stress release, are incorporated into the constitutive model [\cite{bharali2022robust}]. Modeling fracture using the phase field method involves the integration of two fields, namely the vector-valued elastic field, $\boldsymbol{u}(\boldsymbol x)$, and the scalar-valued phase field, $\phi\left(\boldsymbol x\right) \in [0,1]$, with 0 representing the undamaged state of the material and 1 a fully damaged state.

The equilibrium equation for the elastic field for an isotropic model, considering the evolution of crack, can be written as [\cite{goswami2019adaptive}]:
\begin{equation}\label{eq:degradaed_Disp_eq}
    -\nabla\cdot g(\phi)\boldsymbol{\sigma} = \boldsymbol{f} \text{ on } \Omega,
\end{equation}
where $\boldsymbol{\sigma}$ is the Cauchy stress tensor, $\boldsymbol{f}$ is the body force and $g(\phi)= (1-\phi)^2$ represents the monotonically decreasing stress-degradation function that reduces the stiffness of the bulk material in the fracture zone. The elastic field is constrained by Dirichlet and Neumann boundary conditions: 
\begin{equation}\label{eq:Disp_boundary}
    \begin{aligned}
        g(\phi)\boldsymbol{\sigma}\cdot \boldsymbol{n} &= \boldsymbol{t}_N \text{ on } \partial\Omega_{N}, \\
        \boldsymbol{u} &= \boldsymbol{\overline u} \text{ on } \partial \Omega_{D},\\
    \end{aligned}
\end{equation}
where $\boldsymbol{t}_N$ is the prescribed boundary forces and $\boldsymbol{\overline u}$ is the prescribed displacement for each load step. The Dirichlet and Neumann boundaries are represented by $\partial\Omega_{D}$ and $\partial\Omega_{N}$, respectively. Considering the second-order phase field for a quasi-static setup, the governing equation can be written as:
\begin{equation}\label{eq:phasefield_eq}
       \frac{G_c}{l_0}\phi - G_{c}l_{0}\nabla^{2}\phi = -g'(\phi)H(\boldsymbol{x},t; l_c, y_c) \text{ on } \Omega,
\end{equation}
where $G_c$ is a scalar parameter representing the critical energy release rate of the material, $l_0$ is the length scale parameter, which controls the diffusion of the crack, $H(\boldsymbol{x},t)$ is a local strain-history functional, and $y_c$, $l_c$ represent the position and length of the crack respectively. For sharp crack topology, $l_0 \to 0$ [\cite{Bourdin2008}]. $H(\boldsymbol{x},t)$ contains the maximum positive tensile energy ($\Psi^{+}_{0}$) in the history of deformation of the system. The strain-history functional is employed to initialize the crack on the domain as well as to impose irreversibility conditions on the crack growth [\cite{Miehe2010}]. In this problem, we consider $y_c$, $l_c$ to be random variables with $y_c \sim U[0.3,0.7]$ and $l_c \sim U[0.4,0.6]$, thus, the initial strain function $H(\boldsymbol{x},t=0; l_c, y_c)$ is also random (see the Supplementary Materials for more details). We aim to learn the solution operator $\mathcal{G}: H(\boldsymbol{x},t=0; l_c, y_c) \mapsto \phi(\boldsymbol x,t)$ which maps the initial strain-history function to the crack evolution.

\begin{figure}[ht!]
\begin{center}
\includegraphics[width=0.6\textwidth]{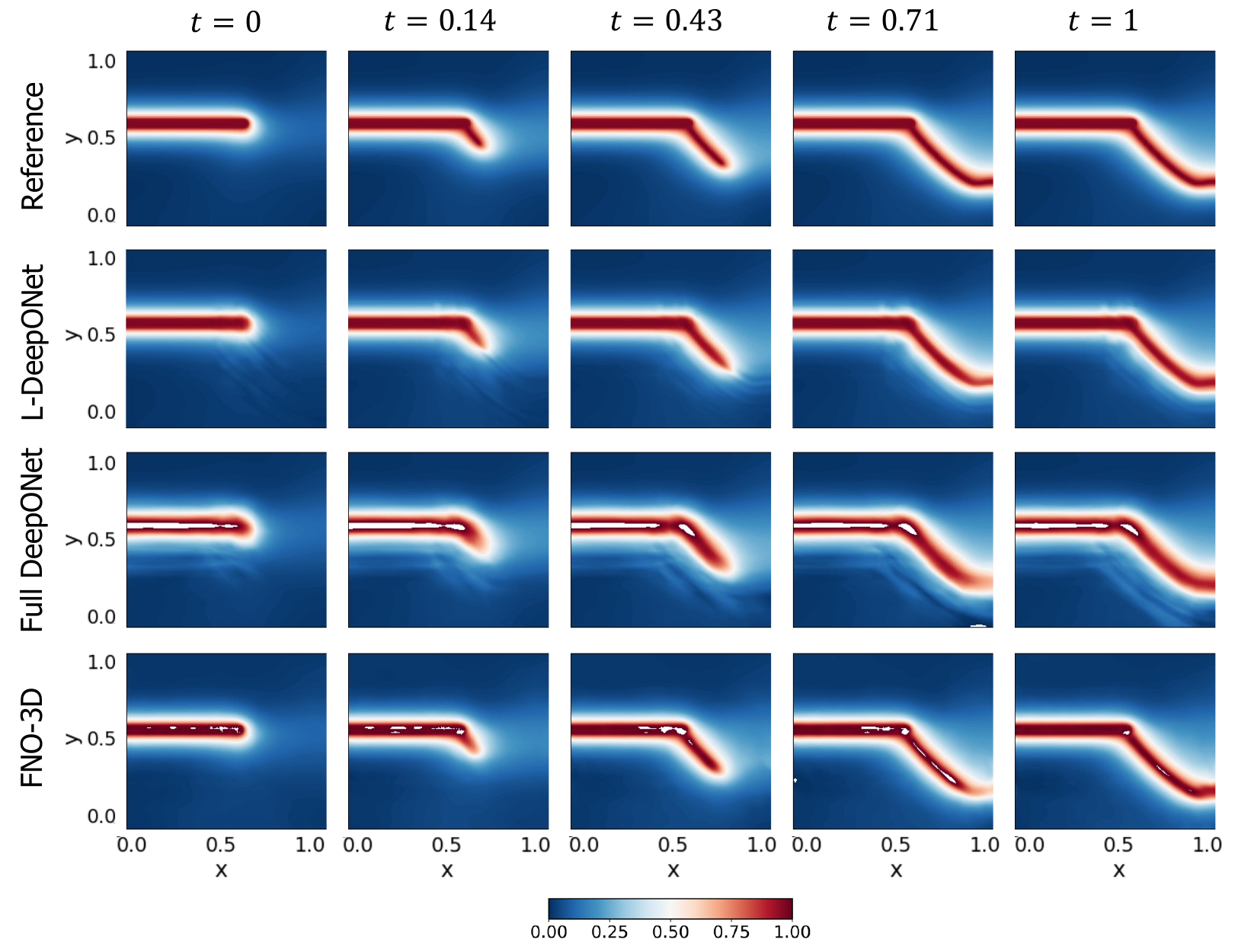}
\caption{Brittle fracture in a plate loaded in shear: results of a representative sample with $y_c = 0.55$ and $l_c = 0.6$ for all neural operators. The results of the L-DeepONet model consider the latent dimension, $d=64$. The neural operator is trained to approximate the growth of the crack for five time steps from a given initial location of the defect.}
\label{fig:fracture}
\end{center}
\captionsetup{justification=centering}
\end{figure}

In Figure \ref{fig:L-DeepONet-results}(a), we show the mean-square error (MSE) between the studied models and ground truth. The left panel shows the MSE for the multi-layer autoencoder (MLAE) for different latent dimensions ($d$), where the violin plot shows the distribution of MSE from $n=5$ independent trials. The right panel shows the resulting MSE for L-DeepONet operating on different latent dimensions ($d$) compared with the full high-dimensional DeepONet, FNO-2D, and FNO-3D. We observe that, regardless of the latent dimension, the L-DeepONet outperforms the standard DeepONet (Full DON) and performs comparably with FNO-2D and FNO-3D. In Figure \ref{fig:fracture}, a comparison between all models for a random representative result is shown. While L-DeepONet results in prediction fields almost identical to the reference, the predictions of the standard models deviate from the ground truth both inside and around the propagated crack. Finally, the cost of training the different models is presented in Table \ref{table:comp_time}. Because the required network complexity is significantly reduced, the L-DeepONet is $1-2$ orders of magnitude cheaper to train than the standard approaches.

\bigbreak
\noindent
\textbf{Rayleigh-Bénard fluid flow convection}

\noindent
Rayleigh-Bénard convection occurs in a thin layer of fluid that is heated from below [\cite{chilla2012new}]. The natural fluid convection is buoyancy-driven and caused due to a temperature gradient $\Delta T$. Instability in the fluid occurs when $\Delta T$ is large enough to make the non-dimensional Rayleigh number, $\text{Ra}$, exceed a certain threshold. The Rayleigh number whose physical interpretation is the ratio between the buoyancy and the viscous forces is defined as
\begin{equation}
    \text{Ra} = \frac{\alpha \Delta T g h^3}{\nu \kappa},
    \label{eq:rayleigh}
\end{equation}
where $\alpha$ is the thermal expansion coefficient, $g$ is the gravitational acceleration, $h$ is the thickness of the fluid layer, $\nu$ is the kinematic viscosity and $\kappa$ is the thermal diffusivity. When $\Delta T$ is small, the convective flow does not occur due to stabilizing effects of viscous friction. Based on the governing conservation laws for an incompressible fluid (mass, momentum, energy) and the Boussinesq approximation according to which density perturbations affect only the gravitational force, the dimensional form of the Rayleigh-Bénard equations for a fluid defined on a domain $\Omega$ reads:
\begin{equation}
\begin{cases}
\displaystyle 
    \frac{\text{D}\boldsymbol u}{\text{D}t} =
    -\frac{1}{\rho_0} \nabla p + \frac{\rho}{\rho_0} g + \nu \nabla^2 \boldsymbol u, & \quad \boldsymbol x \in \Omega, t>0, \\
    \displaystyle 
    \frac{\text{D} T}{\text{D}t} = \kappa \nabla^2 T,  & \quad \boldsymbol x \in \Omega, t>0, \\
    \nabla \cdot \boldsymbol u = 0, \\
    \rho = \rho_0(1-\alpha(T-T_0)),
\end{cases}
\label{eq:R-B}
\end{equation}
where $\text{D} / \text{D}\boldsymbol t$ denotes material derivative, $\boldsymbol u, p, T$ are the fluid velocity, pressure and temperature respectively, $T_0$ is the temperature at the lower plate, and $\boldsymbol x=(x,y)$ are the spatial coordinates. Considering two plates (upper and lower) the corresponding BCs and ICs are defined as 
\begin{equation}
\begin{cases}
\displaystyle 
    T(\boldsymbol x, t) \vert_{y=0} = T_0, & \quad \boldsymbol x \in \Omega, t>0, \\ 
     T(\boldsymbol x, t) \vert_{y=h} = T_1, & \quad \boldsymbol x \in \Omega, t>0, \\ 
    \boldsymbol u(\boldsymbol x,t)\vert_{y=0} = \boldsymbol u(\boldsymbol x,t)\vert_{y=h} = 0,  & \quad \boldsymbol x \in \Omega, t>0, \\
    T(y, t) \vert_{t=0} = T_0 + \frac{y}{h}(T_1 - T_0) + 0.1 v(\boldsymbol x), & \quad \boldsymbol x \in \Omega, \\
     \boldsymbol u(\boldsymbol x,t)\vert_{t=0} = 0, & \quad \boldsymbol x \in \Omega,
\end{cases}
\label{eq:R-B-BCs-ICs}
\end{equation}
where $T_0$, and $T_1$ are the fixed temperatures of the lower and upper plates, respectively. For a 2D rectangular domain and through a non-dimensionalization of the above equations, the fixed temperatures become $T_0=0$ and $T_1=1$. The IC of the temperature field is modeled as linearly distributed with the addition of a GRF, $v(\boldsymbol x)$ having correlation length scales $\ell_x=0.45, \ell_y=0.4$ simulated using a Karhunen-Lo\'eve expansion. The objective is to approximate the operator $\mathcal{G}: T(\boldsymbol x, t=0) \mapsto T(\boldsymbol x, t) $ (see the Supplementary Materials for more details). 

\begin{figure}[ht!]
\begin{center}
\includegraphics[width=1\textwidth]{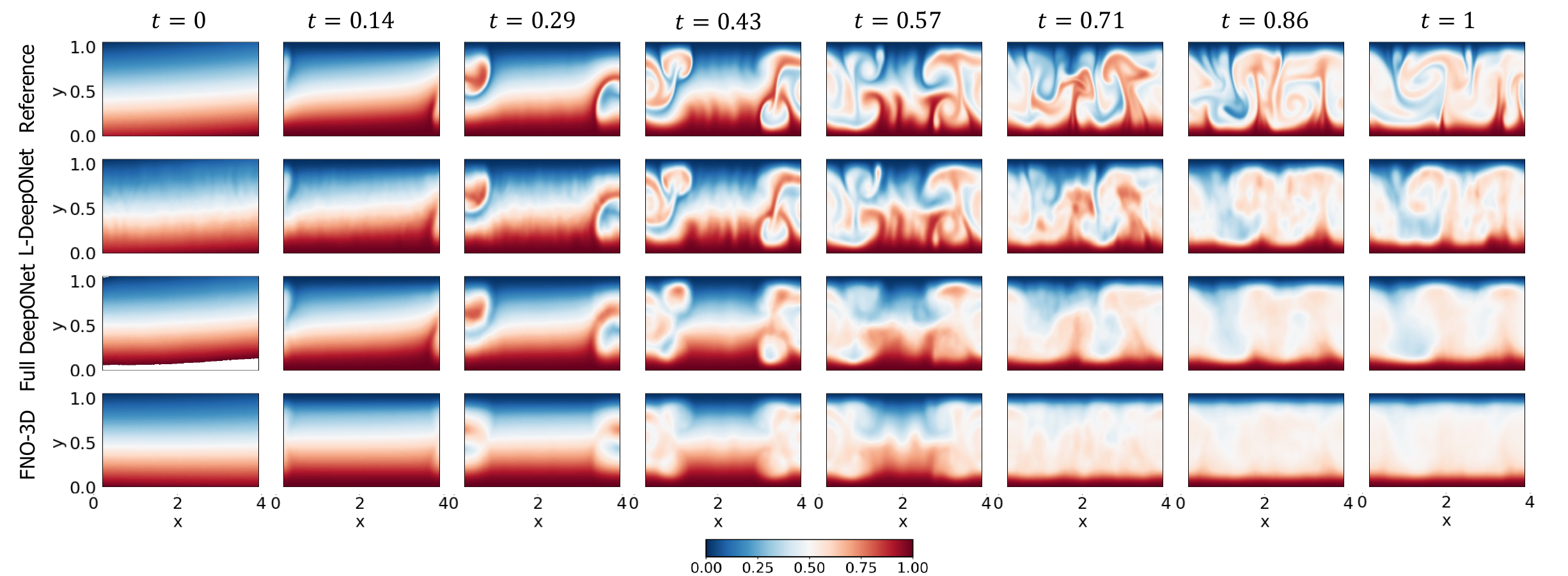}
\caption{Rayleigh-Bénard convective flow: results of the temperature field of a representative sample for all neural operators. The results of the L-DeepONet model consider the latent dimension, $d=100$. The neural operator is trained to approximate the growth of the evolution of the temperature field from a realization of the initial temperature field for seven time steps.}
\label{fig:benard}
\end{center}
\captionsetup{justification=centering}
\end{figure}

Figure \ref{fig:L-DeepONet-results}(b) again shows violin plots of the MSE for the MLAE with differing latent dimensions and the MLE for the corresponding L-DeepONet compared with the other neural operators. Here we see that the reconstruction accuracy of the MLAE is improved by increasing the latent dimensionality up to $d=100$. However, the change in the predictive accuracy of L-DeepONet for different values of $d$ is less significant, indicating that latent spaces with even very small dimensions ($d=25$) result in a very good performance. Furthermore, L-DeepONet outperforms all other neural operators with a particularly significant improvement compared to FNO. In Figure \ref{fig:benard}, we observe that L-DeepONet is able to capture the complex dynamical features of the true model with high accuracy as the simulation evolves. In contrast, the standard DeepONet and FNO result in diminished performance as they tend to smooth out the complex features of the true temperature fields. Furthermore, the training time of the L-DeepONet is significantly lower than the full DeepONet and FNO as shown in Table \ref{table:comp_time}.

\noindent

\bigbreak
\noindent
\textbf{Shallow-water equations}

\noindent
The shallow-water equations model the dynamics of large-scale atmospheric flows [\cite{galewsky2004initial}]. In a vector form, the viscous shallow-water equations can be expressed as
\begin{equation}
\begin{cases}
\displaystyle 
    \frac{\text{D}\boldsymbol V}{\text{D}t} = -f \boldsymbol k \times \boldsymbol V -g \nabla h + \nu \nabla^2 \boldsymbol V, \\[1ex]
    \displaystyle 
    \frac{\text{D} h}{\text{D}t} = -h \nabla \cdot \boldsymbol V + \nu \nabla^2 h, \quad \boldsymbol x \in \Omega, \ t \in [0,1],
\end{cases}
\label{eq:shallow-water}
\end{equation}
where $\Omega= ( \lambda, \phi)$ represents a spherical domain where $\lambda, \phi$ are the longitude and latitude respectively ranging from $[-\pi,\pi]$,  $\boldsymbol V = \boldsymbol i u + \boldsymbol j v $ is the velocity vector tangent to the spherical surface ($\boldsymbol i$ and $\boldsymbol j$ are the unit vectors in the eastward and northward directions respectively and $u, v$ the velocity components), and $h$ is the height field which represents the thickness of the fluid layer. Moreover, $f = 2 \Xi \sin \phi$ is the Coriolis parameter, where $\Xi$ is the Earth's angular velocity, $g$ is the gravitational acceleration and $\nu$ is the diffusion coefficient. 

As an initial condition, we consider a zonal flow which represents a typical mid-latitude tropospheric jet. The initial velocity component $u$ is expressed as a function of the latitude $\phi$ as
\begin{equation}
u(\phi, t=0)=
\begin{cases}
\displaystyle 
   \hspace{60pt} 0
    & \quad \text{for} \quad \phi \le \phi_0, \\
    \displaystyle 
    \frac{u_{\text{max}}}{n} \text{exp} \Bigg[ \frac{1}{(\phi - \phi_0)(\phi - \phi_1)} \Bigg]& \quad \text{for} \quad \phi_0 < \phi < \phi_1, \\
    \hspace{60pt} 0 & \quad \text{for} \quad \phi \ge \phi_1,
\end{cases}
\label{eq:ICs-shallow-water}
\end{equation}
where $u_{\text{max}}$ is the maximum zonal velocity, $\phi_0$, and $\phi_1 $ represent the latitude in the southern and northern boundary of the jet in radians, respectively, and $n=\text{exp}[-4/(\phi_1 - \phi_0)^2]$ is a non-dimensional parameter that sets the value $u_{\text{max}}$ at the jet's mid-point. A small unbalanced perturbation is added to the height field to induce the development of barotropic instability. The localized Gaussian perturbation is described as
\begin{equation}
    h'(\lambda, \phi, t=0) = \hat{h} \cos (\phi) \exp[-(\lambda/ \alpha)^2] \exp[-(\phi_2-\phi)/\beta]^2, 
    \label{eq:perturbation}
\end{equation}
where $-\pi < \lambda < \pi$ and $\hat{h}, \phi_2, \alpha, \beta$ are parameters that control the location and shape of the perturbation. We consider $\alpha, \beta$ to be random  variables with $\alpha \sim U[0.\bar{1},0.5]$ and $\beta \sim U[0.0\bar{3},0.2]$ so that the input Gaussian perturbation is random. The localized perturbation is added to the initial height field, which forms the final initial condition $h(\lambda, \phi, t=0)$ (see Supplementary Materials for more details). The objective is to approximate the operator $\mathcal{G}:h(\lambda, \phi, t=0) \mapsto u(\lambda, \phi, t)$. This problem is  particularly challenging as the fine mesh required to capture the details of the convective flow both spatially and temporally results in output realizations having millions of dimensions. 

\begin{figure}[ht!]
\begin{center}
\includegraphics[width=0.9\textwidth]{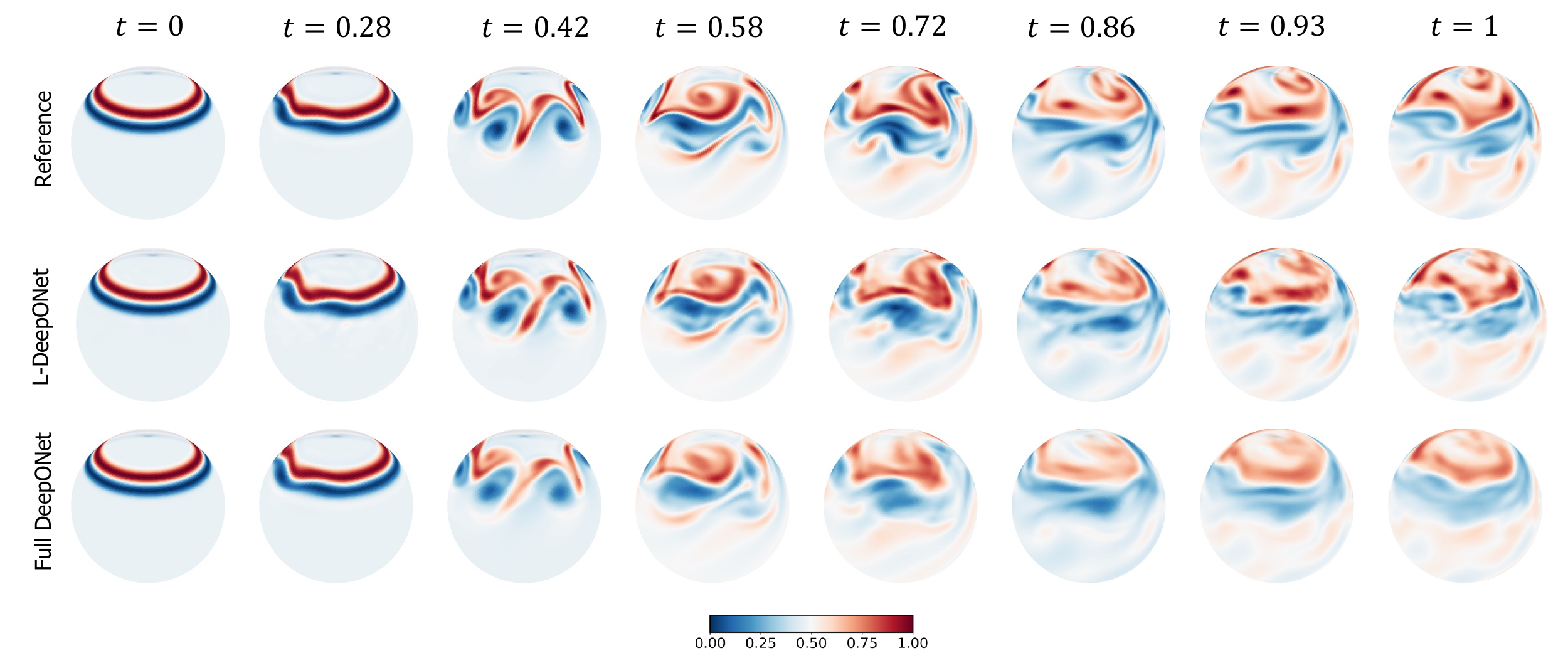}
\caption{Shallow water equations: results of the evolution of the velocity field through eight time steps for all the operator models considered in this work, for a representative realization of the initial perturbation to the height field. The results of the L-DeepONet model consider the latent dimension, $d=81$.}
\label{fig:shallow}
\end{center}
\captionsetup{justification=centering}
\end{figure}

Unlike the previous two applications, here the approximated operator learns to map the initial condition of one quantity, $h(\lambda, \phi, t=0)$, to the evolution of a different quantity, $u(\lambda, \phi, t)$. Given the difference between the input and output quantities of interest (in scale and features), a single encoding of the combined data as in the standard proposed approach (see Figure \ref{fig:L-DeepONet}) is insufficient. Instead, two separate encodings are needed for the input and output data, respectively. While an autoencoder is used to reduce the dimensionality of the output data representing the longitudinal component of the velocity vector $u$, standard principal component analysis (PCA) is performed on the input data due to the small local variations in the initial random height field $h$ which results in a small intrinsic dimensionality. 

Results, in terms of MSE, are presented in Figure \ref{fig:L-DeepONet-results}(c), where again we see that the L-DeepONet outperforms the standard approach while changes in the latent dimension do not result in significant differences in the model accuracy. Consistent with the results of the previous application, the training cost of the L-DeepONet is much lower than the full DeepONet (Table \ref{table:comp_time}). We further note that training FNO for this problem (either FNO-2D or FNO-3D) proved computationally prohibitive. For a moderate 3D problem with spatial discretization beyond $64^3$, the latest GPU architectures such as the NVIDIA Ampere GPU do not provide sufficient memory to process a single training sample [\cite{grady2022towards}]. Data partitioning across multiple GPUs with distributed memory, model partitioning techniques like pipeline parallelism, and domain decomposition approaches [\cite{grady2022towards}] can be implemented to handle high-dimensional tensors within the context of an automatic differentiation framework to compute the gradients/sensitivities of PDEs and thus optimize the network parameters. This advanced implementation is beyond the scope of this work as it proves unnecessary for the studied approach. Consequently, a comparison to the FNO is not shown here. Figure \ref{fig:shallow}, shows the evolution of the L-DeepONet and the full DeepONet compared to the ground truth for a single realization. The L-DeepONet consistently captures the complex nonlinear dynamical features for all time steps, while the full model prediction degrades over time and again smoothing the results such that it fails to predict extreme velocity values for each time step that can be crucial, e.g., in weather forecasting. 


\begin{table}[ht!]
\begin{center}
\caption{Comparison of the computational training time in seconds (s) for all the neural operators across all considered applications, identically trained on an NVIDIA A6000 GPU. Inference is performed at a fraction of a second for all the approaches.}
\label{table:comp_time}
\begin{tabular}{ l c c c c } 
\hline
Application & L-DeepONet & Full DeepONet & FNO-3D  \\
\hline
Brittle material fracture & $1\small{,}660$ & $15\small{,}031$ & $128\small{,}000$ \\ 
Rayleigh-Bénard fluid flow & $2\small{,}853$ & $6\small{,}772$ & $1\small{,}126\small{,}400$   \\ 
Shallow water equation & $15\small{,}218$ & $379\small{,}022$ & --  \\ 
\hline
\end{tabular}
\label{table:computational_time-MLAE}
\end{center}
\end{table}

\begin{table}[ht!]
\begin{center}
\caption{Comparison of the accuracy of the L-DeepONet for two different dimensionality reduction techniques; namely, the multi-layer autoencoders (MLAE) and principal component analysis (PCA), and $d$ denotes the size of the latent space. Results for both the maximum and minimum $d$ values tested for each applications are provided. To evaluate the performance of L-DeepONet, we compute the mean square error of predictions, and we report the mean and standard deviation of this metric based on five independent training trials.}
\begin{tabular}{ l c c c c } 
\hline
Application & $d$ & with MLAE & with PCA  \\
\hline
\multirow{2}{*}{Brittle material fracture} & $9$ & $3.33 \cdot 10^{-4} \pm 4.99 \cdot 10^{-5}$ & $2.71 \cdot 10^{-3} \pm 6.62 \cdot 10^{-6}$ \\ 
& $64$ & $2.02 \cdot 10^{-4} \pm 1.88 \cdot 10^{-5}$ & $3.13 \cdot 10^{-4} \pm 4.62 \cdot 10^{-6}$ \\ \hline
\multirow{2}{*}{Rayleigh-Bénard fluid flow} & $25$ & $4.10 \cdot 10^{-3} \pm 8.05 \cdot 10^{-5}$ & $3.90 \cdot 10^{-3} \pm 4.73 \cdot 10^{-5}$  \\ & $100$ & $3.55 \cdot 10^{-3} \pm 1.46 \cdot 10^{-4}$ & $3.76 \cdot 10^{-3} \pm 4.86 \cdot 10^{-5}$  \\ \hline
\multirow{2}{*}{Shallow water equation} & $25$ & $2.30 \cdot 10^{-4} \pm 1.50 \cdot 10^{-5}$ & $7.98 \cdot 10^{-4} \pm 8.01 \cdot 10^{-7}$ \\ & $81$ & $2.23 \cdot 10^{-4} \pm 1.83 \cdot 10^{-5}$ & $4.18 \cdot 10^{-4} \pm 4.67 \cdot 10^{-6}$ \\
\hline
\end{tabular}
\label{table:accuracy_mlae_vs_pca}
\end{center}
\end{table}

\section{Discussion}

We have investigated latent DeepONet (L-DeepONet) for learning neural operators on latent spaces for time-dependent PDEs exhibiting highly non-linear features both spatially and temporally and resulting in high-dimensional observations. The L-DeepONet framework leverages autoencoder models to cleverly construct compact representations of the high-dimensional data while a neural operator is trained on the identified latent space for operator regression. Both the advantages and limitations of L-DeepONet are demonstrated on a collection of diverse PDE applications of increasing complexity and data dimensionality. As presented, L-DeepONet provides a powerful tool in SciML and UQ that improve the accuracy and generalizability of neural operators in applications where high-fidelity simulations are considered to exhibit complex dynamical features, e.g., in climate models. 

A systematic comparison with standard DeepONet and FNO revealed that L-DeepONet improves the quality of results and it can capture with greater accuracy the evolution of the system represented by a time-dependent PDE. This result is more noticeable as the dimensionality and non-linearity of dynamical features increase (e.g., in complex convective fluid flows). Another advantage is that L-DeepONet training requires less computational resources, as standard DeepONet and FNO are trained on the full-dimensional data and are thus, more computationally demanding and require much larger memory (see Table \ref{table:comp_time}). For all applications, we found that a small latent dimensionality ($d\le 100)$ is sufficient for constructing powerful neural operators, by removing redundant features that can hinder the network optimization and thus its predictive accuracy. Furthermore, L-DeepONet can alleviate the computational demand and thus enable  tasks that require the computation of kernel matrices, e.g., used in transfer learning for comparing the statistical distance between data distributions [\cite{goswami2022deep}]. 

Despite the advantages of learning operators in latent spaces, there are certain limitations that warrant discussion. L-DeepONet trains DR models to identify suitable latent representations for the combined input and output data. However, as shown in the final application, in cases where the approximated mapping involves heterogeneous quantities, two independent DR models need to be constructed. 
While in this work we found that simple MLAE models result in the smallest L-DeepONet predictive error, a preliminary study regarding the suitability of the DR approach needs to be performed for all quantities of interest. 
Another disadvantage is that the L-DeepONet as formulated is unable to interpolate in the spatial dimensions. The current L-DeepONet consists of a modified trunk net where the time component has been preserved while the spatial dimensions have been convolved. Thus, L-DeepONet can be used for time but not for space interpolation/extrapolation. Finally, L-DeepONet cannot be readily employed in a physics-informed learning manner since the governing equations are not known in the latent space and therefore cannot be directly imposed. These limitations motivate future studies that continue to assist researchers in the process of constructing accurate and generalizable surrogate models for complex PDE problems prevalent in physics and engineering. 

\section{Materials and Methods}

\subsection{Problem statement}

Neural operators learn nonlinear mappings between infinite dimensional functional spaces on bounded domains and provide a unique simulation framework for real-time inference of complex parametric PDEs. Let $\Omega \subset \mathbb{R}^D$ be a bounded open set and $\mathcal{X}=\mathcal{X}(\Omega; \mathbb{R}^{d_x})$ and $\mathcal{Y}=\mathcal{Y}(\Omega;\mathbb{R}^{d_y})$ two separable Banach spaces. Furthermore, assume that $\mathcal{G}: \mathcal{X} \rightarrow \mathcal{Y}$ is a non-linear map arising from the solution of a time-dependent PDE. The objective is to approximate the nonlinear operator via the following parametric mapping
\begin{equation}
\begin{aligned}
    \mathcal{G}: \mathcal{X} \times \Theta \rightarrow \mathcal{Y} \hspace{15pt} \text{or}, \hspace{15pt} \mathcal{G}_{\theta}: \mathcal{X} \rightarrow \mathcal{Y}, \hspace{5pt} \theta \in \Theta
\end{aligned}
\label{eq:PDE-approx}
\end{equation}
where $\Theta$ is a finite-dimensional parameter space. In this standard setting, the optimal parameters $\theta^*$ are learned through training the neural operator (e.g., via DeepONet, FNO) with a set of labeled observations $\{\mathbf{x}_j, \mathbf{y}_j \}_{j=1}^N$ generated on a discretized domain $\Omega_m = \{x_1, \dots, x_m\} \subset \Omega$ where $\{x_j\}_{j=1}^m$ represent the sensor locations, thus $\mathbf{x}_{j|\Omega_m} \in \mathbb{R}^{D_x}$ and $\mathbf{y}_{j|\Omega_m} \in \mathbb{R}^{D_y}$ where $D_x= d_x \times m$ and $D_y = d_y \times m$. Representing the domain discretization with a single parameter $m$, corresponds to the simplistic case where mesh points are equispaced. However, the training data of neural operators are not restricted to equispaced meshes. For example, for a time-dependent PDE with two spatial and one temporal dimension with discretizations $m_s, m_t$ respectively, the total output dimensionality is computed as $D_y= m_s^{d_x} \times m_t$.

\subsection{Approximating nonlinear operators on latent spaces via L-DeepONet}

In physics and engineering, we often consider high-fidelity time-dependent PDEs generating very high-dimensional input/output data with complex dynamical features. To address the issue of high dimensionality and improve the predictive accuracy we employ L-DeepONet which allows the training of DeepONet on latent spaces. The approach involves two main steps: 1) the nonlinear DR of both input and output data $\{\mathbf{x}_j, \mathbf{y}_j \}_{j=1}^N$ via a suitable and invertible DR technique, 2) learning of a DeepONet model on a latent space and inverse transformation of predicted samples back to the original space. This process is defined as
\begin{equation}
\begin{aligned}
     &\mathcal{J}_{\theta_{\text{encoder}}} \colon \{\mathbf{x}, \mathbf{y}\} \mapsto \{\mathbf{x}^r, \mathbf{y}^r\} \\
    &\mathcal{G}_{\theta} \colon \mathbf{x}^r \mapsto \mathbf{y}^r \\
    &\mathcal{J}_{\theta_{\text{decoder}}} \colon \mathbf{y}^r \mapsto \mathbf{y}^{\text{rec}} 
\end{aligned}
\label{eq:l-deeponet}
\end{equation}
where $\mathcal{J}_{\theta_{\text{encoder}}}, \mathcal{J}_{\theta_{\text{decoder}}}$ are the two parts of a DR method, $r$ corresponds to data on the reduced space, $\mathcal{G}_{\theta}$ is the approximated latent operator and $\theta$ its trainable parameters. While the encoder $\mathcal{J}_{\theta_{\text{encoder}}}$ is used to project high-dimensional data onto the latent space, the decoder $\mathcal{J}_{\theta_{\text{decoder}}}$ is employed during the training of DeepONet to project predicted samples back to original space and evaluate its accuracy on the full-dimensional data $\{\mathbf{x}_j, \mathbf{y}_j \}_{j=1}^N$. Once trained, L-DeepONet can be used for real-time inference at no cost. We note that the term `L-DeepONet' refers to the trained DeepONet model together with the pre-trained encoder and decoder parts of the autoencoder which are required to perform inference in unseen samples (see Figure \ref{fig:L-DeepONet}). Next, the distinct parts of the L-DeepONet framework are elucidated in detail.

\subsubsection*{Learning latent representations}

\noindent
The first objective is to identify a latent representation for the high-dimensional input/output PDE data. Compressing the data to a reduced representation will not only allow us to accelerate the DeepONet training but, as shown above, it improves predictive performance and robustness. To this end, we employ autoencoders due to their flexibility in the choice of the model architecture and the inherent inverse mapping capability. We note that the proposed framework allows for the adoption of any suitable linear or nonlinear DR method provided the existence of an inverse mapping. In this work, the objective is to demonstrate that DR enhances the accuracy of neural operators rather than establishing which DR method is the most advantageous. The latter depends on various factors including accuracy, generalizability, and computational cost. For our demonstrations, we apply AEs that we found to perform comparably or better than PCA across our diverse set of PDEs through systematic study (see Table \ref{table:accuracy_mlae_vs_pca} and Supplementary Materials). However, the choice of DR approach can be problem and resource-dependent so, although AEs generally outperform PCA, PCA is found to be a viable approach for many problems and under certain conditions.

We train unsupervised autoencoder model $\mathcal{J}_{\theta_{\text{ae}}}$ and perform hyperparameter tuning to identify the optimal latent dimensionality $d$, where $d \ll D_x, D_y$. Assume a time-dependent PDE, where $d_x$ corresponds to the dimensionality of the input space and $m_s, m_t$ the spatial and temporal discretizations of the generated data. In order to feed the autoencoder model with image-like data, the PDE outputs are reshaped into distinct snapshots, i.e., $\{\hat{\mathbf{y}_i}\}_{i=1}^{N\times m_t}$. Finally, input and output data are concatenated into a single dataset  $\{\mathbf{z}_i\}_{i=1}^{N(1+m_t)}$. The two parts of the autoencoder model, which are trained concurrently, are expressed as
\begin{equation}
\begin{aligned}
    \mathcal{J}_{\theta_{\text{encoder}}} \colon \{\mathbf{x}, \hat{\mathbf{y}}\} \equiv \mathbf{z} \mapsto \{\mathbf{x}^r, \mathbf{y}^r\} \equiv \mathbf{z}^r, \\
    \mathcal{J}_{\theta_{\text{decoder}}} \colon \{\mathbf{x}^r, \mathbf{y}^r\} \equiv \mathbf{z}^r  \mapsto \{\tilde{\mathbf{x}}, \tilde{\mathbf{y}}\} \equiv \tilde{\mathbf{z}} ,
\label{eq:ae-mappings}
\end{aligned}
\end{equation}
where $\{\mathbf{x}^r_i\}_{i=1}^N \in \mathbb{R}^d$, $\{\mathbf{y}^r_i\}_{i=1}^{N \times m_t} \in \mathbb{R}^d$ and $\{\mathbf{z}^r_i\}_{i=1}^{N(1+m_t)} \in \mathbb{R}^d$. The trainable parameters of the encoder and decoder are represented with $\theta_{\text{encoder}}$ and $\theta_{\text{decoder}}$ respectively. The optimal set of the autoencoder parameters $\theta_{\text{ae}}= \{\theta_{\text{encoder}}, \theta_{\text{decoder}} \}$ are obtained via the minimization of the loss function
\begin{equation}
    \mathcal{L}_{\text{ae}} = \min_{\theta_{\text{ae}}} \| \mathbf{z} - \mathbf{\tilde{z}} \|^2_2,
\label{eq:ae-loss}
\end{equation}
where ${\|\cdot\|}_2$ denotes the standard Euclidean norm and $\tilde{\mathbf{z}}\ \equiv \{\tilde{\mathbf{x}},  \tilde{\mathbf{y}} \}$ denotes the reconstructed dataset of combined input and output data. From a preliminary study, which is not shown here for the sake of brevity, we investigated three AE models, simple autoencoders (vanilla-AE) with a single hidden layer, multi-layer autoencoders (MLAE), with multiple hidden layers and convolutional autoencoders (CAE) which convolve data through convolutional layers.  We found that MLAE performs best, even with a small number of hidden layers (e.g., $3$). Furthermore, the use of alternative AE models which are primarily used as generative models, such as variational autoencoders (VAE) [\cite{kingma2013auto}] or Wasserstein autoencoders (WAE) [\cite{tolstikhin2017wasserstein}], resulted in significantly worse L-DeepONet performance. Although such models resulted in good reconstruction accuracy and thus can be used to reduce the data dimensionality and generate synthetic yet realistic samples, we found that the obtained submanifold is not well-suited for training the neural operator, as it may result in the reduction of data variability or even representation collapse. 

\subsubsection*{Training neural operator on latent space (L-DeepONet)}

\noindent
Once the autoencoder model is trained and the reduced data $\{\mathbf{x}^r, \mathbf{y}^r\} $ are generated, we aim to approximate the latent representation mapping with an unstacked DeepONet $\mathcal{G}_{\theta}$, where $\theta$ are the trainable model parameters. As shown in Figure \ref{fig:L-DeepONet}, the unstacked DeepONet consists of two concurrent DNNs, a branch net which encodes the inputs realizations $\mathbf{x}^r \in \mathbb{R}^d$ (in this case the reduced input data) evaluated at the reduced spatial locations $\{x_1, x_2, \dots ,x_d\}$. On the other hand, the trunk net takes as input the temporal coordinates $\zeta= \{ t_i \}_{i=1}^{m_t}$ at which the PDE output is evaluated. The solution operator for an input realization, $\mathbf x_1$, can be expressed as:
\begin{equation}\label{eq:output_deeponets}
    \begin{aligned}
      \mathcal G_{\theta}(\mathbf{x}^r_1)(\zeta) &= \sum_{i = 1}^p b_i \cdot tr_i = \sum_{i = 1}^{p}b_i(\mathbf{x}^r_{1}(x_1), \mathbf{x}^r_{1}(x_2), \ldots, \mathbf {x}^r_{1}(x_d))\cdot tr_i(\zeta),   
    \end{aligned}
\end{equation}
where $[b_1, b_2, \ldots, b_p]^T$ is the output vector of the branch net, $[tr_1, tr_2, \ldots, tr_p]^T$ the output vector of the trunk net and $p$ denotes a hyperparameter that controls the size of the final hidden layer of both the branch and trunk net. The trainable parameters of the DeepONet, represented by $\theta$ in Eq.~\eqref{eq:output_deeponets}, are obtained by minimizing a loss function, which is expressed as:
\begin{equation}
\begin{aligned}
    \mathcal L(\theta) &= \mathcal L_r(\theta) + \mathcal L_i(\theta), \\
    \mathcal{L}_r(\theta) &=  \min_{\theta} \| \mathbf{y}^r - \tilde{\mathbf{y}}^r \|^2_2,
\end{aligned}
\end{equation}
where $\mathcal L_r(\theta)$, $\mathcal L_i(\theta)$ denote the residual loss and the initial condition loss respectively, $\mathbf{y}^r$ the reference reduced outputs and $\tilde{\mathbf{y}}^r$ the predicted reduced outputs. In this work, we only consider the standard regression loss $\mathcal{L}_r(\theta)$, however, additional loss terms can be added to the loss function. The branch and trunk networks can be modeled with any specific architecture. Here we consider a CNN for the branch net architecture and a feed-forward neural network (FNN) for the trunk net to take advantage of the low dimensions of the evaluation points, $\zeta$. To feed the branch net of L-DeepONet the reduced output data are reshaped to $\mathbb{R}^{\sqrt{d}\times\sqrt{d}}$, thus it is advised to choose square latent dimensionality values. Once the optimal parameters $\theta$ are obtained, the trained model can be used to predict the reduced output for novel realizations of the input $\mathbf{x} \in \mathbb{R}^{D_{x}}$. Finally, the predicted data are used as inputs to the pre-trained decoder $\mathcal{J}_{\theta_{\text{decoder}}}$, to transform results back to the original space and obtain the approximated full-dimensional output  $\mathbf{y}^{\text{rec}} \in \mathbb{R}^{D_{y}}$. We note that the training cost of L-DeepONet is significantly lower compared to the standard model, due to the smaller size of the network and the reduced total number of its trainable parameters.

\subsubsection*{Error metric}

\noindent
To assess the performance of L-DeepONet we consider the MSE evaluated on a set of $N_{\text{test}}$ test realizations
\begin{equation}
\label{eq:mse}
    \text{MSE} = \frac{1}{N_{\text{test}}} \sum_{i=1}^{N_{\text{test}}}\big(\mathbf{y}_i - \mathbf{y}_i^{\text{rec}} \big)^2,
\end{equation}
where $\mathbf{y}\in \mathbb{R}^{D_{y}}$ is the reference and $\mathbf{y}^{\text{rec}}\in \mathbb{R}^{D_{y}}$ the predicted output respectively.

More details on how this framework is implemented for different PDE systems of varying complexity can be found in Results (Section~\ref{sec:results}). Information regarding the choice of neural network architectures and generation of training data are provided in the Supplementary Materials.

\bibliography{mybibfile}

\section*{Acknowledgements}
The authors would like to acknowledge computing support provided by the Advanced Research Computing at Hopkins (ARCH) core facility at Johns Hopkins University and the Rockfish cluster and the computational resources and services at the Center for Computation and Visualization (CCV), Brown University where all experiments were carried out. 

\subsection*{Funding}
KK \& MDS: U.S. Department of Energy, Office of Science, Office of Advanced Scientific Computing Research grant under Award Number DE-SC0020428. \\
SG \& GEK: U.S. Department of Energy project PhILMs under Award Number DE-SC0019453 and the OSD/AFOSR Multidisciplinary Research Program of the University Research Initiative (MURI) grant FA9550-20-1-0358.

\subsection*{Author contributions}
Conceptualization: KK, SG, GEK, MDS \\
Investigation: KK, SG \\
Visualization: KK, SG \\
Supervision: GEK, MDS \\
Writing—original draft: KK, SG \\
Writing—review \& editing: KK, SG, GEK, MDS

\subsection*{Data and materials availability}
All data needed to evaluate the conclusions in the paper are presented in the paper and/or the Supplementary Materials. All code and data accompanying this manuscript will become publicly available at \url{https://github.com/katiana22/latent-deeponet} upon publication of the paper.


\subsection*{Competing interests}
The authors declare no competing interests.

\makeatletter
\newcommand*{\addFileDependency}[1]{
  \typeout{(#1)}
  \@addtofilelist{#1}
  \IfFileExists{#1}{}{\typeout{No file #1.}}
}
\makeatother

\newcommand*{\myexternaldocument}[1]{%
    \externaldocument{#1}%
    \addFileDependency{#1.tex}%
    \addFileDependency{#1.aux}%
}



\newpage
\vspace*{10\baselineskip}
\centerline{\LARGE{Supplementary Materials for}}
\vspace*{2\baselineskip}
\centerline{\Large{\textbf{Learning in latent spaces improves the predictive accuracy}}}
\centerline{\Large{\textbf{of deep neural operators}}}
\vspace{20pt}
\centerline{Kontolati Katiana, Goswami Somdatta, George Em Karniadakis, Michael D Shields*}
\vspace{20pt}
\centerline{*Corresponding author. Email: \href{mailto:michael.shields@jhu.edu}{michael.shields@jhu.edu}}
\vspace{50pt}
\noindent\textbf{\large{This PDF file includes:}}\vspace{5pt} \\
\indent\large{Supplementary Text} \\
\indent\large{Tables S1 to S3} \\
\indent\large{Figures S1 to S7} \\
\indent\large{References}

\makeatletter
\renewcommand \thesection{S\@arabic\c@section}
\renewcommand\thetable{S\@arabic\c@table}
\renewcommand \thefigure{S\@arabic\c@figure}
\makeatother
\setcounter{figure}{0}
\setcounter{table}{0}
\setcounter{section}{0}
\setcounter{page}{1}

\normalsize
\newpage
\section*{Supplementary Text}

\section*{Nomenclature}
\label{sec:Nomenclature} 

\begin{table}[h!]
\caption{Summary of the main symbols and notation used in this work.}
\centering
\begin{tabular}{c c}
\toprule
 Notation & Description  \\
 \toprule
$\mathbf{x}_j$ & an input realization (e.g., ICs, BCs) \\
$\mathbf{y}_j$ & an output of the PDE model \\
$\boldsymbol f(\cdot)$ & a forcing function of the PDE \\
$\mathcal{G}$ & PDE solution operator  \\
$\mathcal{G}_{\theta}$ & approximation of mapping on latent space  \\
$\theta$ & trainable parameters of the neural operator \\
 $\mathcal{J}_{\theta_{\text{encoder}}}$ & encoder part of the autoencoder \\
  $\mathcal{J}_{\theta_{\text{decoder}}}$ & decoder part of the autoencoder \\
$\{x_i\}_{i=1}^m$ & sensor locations \\
$m_s, m_t$ & spatial and temporal discretization \\
$[\mathbf{x}^r_j(x_1), \mathbf{x}^r_j(x_2),.. \mathbf{x}^r_j(x_d)]$ & pointwise evaluation of the reduced input to
the branch net \\
$\zeta$ & locations as inputs to the trunk net \\
$\mathcal{L}_{\text{ae}}$ & autoencoder loss \\
$\mathcal{L}_r(\theta)$ & L-DeepONet residual loss \\
$d$ & latent space dimensionality \\
\text{GRF} & Gaussian random field \\
\text{CNN} & convolutional neural network \\
\text{FNN} & feed-forward neural network \\
\text{CAE} & convolutional autoencoder \\
\text{VAE} & variational autoencoder \\
\text{MLAE} & multi-layer autoencoder \\
$N$ & total number of train/test data \\
OOD & out-of-distribution \\
KLE & Karhunen-Lo\'eve expansion \\
MSE & mean squared error \\
\bottomrule
\end{tabular}
\label{table:nomenclature}
\end{table}

\section*{Theoretical details}
\label{sec:theoretical-details} 

\subsection*{Neural operators}

Let $\Omega \subset \mathbb{R}^D$ be a bounded open set and $\mathcal{X}=\mathcal{X}(\Omega; \mathbb{R}^{d_x})$ and $\mathcal{Y}=\mathcal{Y}(\Omega;\mathbb{R}^{d_y})$ two separable Banach spaces. Furthermore, assume that $\mathcal{G}: \mathcal{X} \rightarrow \mathcal{Y}$ is a non-linear map arising from the solution of a time-dependent PDE. The objective is to approximate the nonlinear operator via the following parametric mapping
\begin{equation}
\begin{aligned}
    \mathcal{G}: \mathcal{X} \times \Theta \rightarrow \mathcal{Y} \hspace{15pt} \text{or}, \hspace{15pt} \mathcal{G}_{\theta}: \mathcal{X} \rightarrow \mathcal{Y}, \hspace{5pt} \theta \in \Theta
\end{aligned}
\end{equation}
where $\Theta$ is a finite dimensional parameter space. The optimal parameters $\theta^*$ are learned via the training of a neural operator with backpropagation based on a dataset $\{\mathbf{x}_j, \mathbf{y}_j \}_{j=1}^N$ generated on a discretized domain $\Omega_m = \{x_1, \dots, x_m\} \subset \Omega$ where $\{x_j\}_{j=1}^m$ represent the sensor locations, thus $\mathbf{x}_{j|\Omega_m} \in \mathbb{R}^{D_x}$ and $\mathbf{y}_{j|\Omega_m} \in \mathbb{R}^{D_y}$ where $D_x= d_x \times m$ and $D_y = d_y \times m$. 

\subsubsection*{DeepONet}

\newtheorem{theorem}{Theorem}

The Deep Operator Network (DeepONet) [\cite{lu2021learning}] aims to learn operators between infinite-dimensional Banach spaces. Learning is performed in a general setting in the sense that the sensor locations $\{x_i\}_{i=1}^m$ at which the input functions are evaluated need not be equispaced, however they need to be consistent across all input function evaluations. Instead of blindly concatenating the input data (input functions $[\mathbf{x}(x_1), \mathbf{x}(x_2), \dots, \mathbf{x}(x_m)]^T$ and locations $\zeta$)
as one input, i.e., $[\mathbf{x}(x_1), \mathbf{x}(x_2), \dots, \mathbf{x}(x_m), \zeta]^T$, DeepONet employs two subnetworks and treats the two inputs equally. Thus, DeepONet can be applied for high-dimensional problems, where the dimension of $\mathbf{x}(x_i)$ and $\zeta$ no longer match since the latter is a vector of $d$ components in total. A trunk network $\mathbf{f}(\cdot)$, takes as input $\zeta$ and outputs $[tr_1, tr_2, \ldots, tr_p]^T \in \mathbb{R}^p$ while a second network, the branch net $\mathbf{g}(\cdot)$, takes as input $[\mathbf{x}(x_1), \mathbf{x}(x_2), \dots, \mathbf{x}(x_m)]^T$ and outputs $[b_1, b_2, \ldots, b_p]^T \in \mathbb{R}^p$. Both subnetwork outputs are merged through a dot product to generate the quantity of interest. A bias $b_0 \in \mathbb{R}$ is added in the last stage to increase expressivity, i.e., $\mathcal{G}(\mathbf{x})(\zeta) \approx \sum_{i=k}^p b_k t_k + b_0$. The generalized universal approximation theorem for operators, inspired by the original theorem introduced by \cite{chen1995universal}, is presented below. The generalized theorem essentially replaces shallow networks used for the branch and trunk net in the original work with deep neural networks to gain expressivity.

\begin{theorem}[Generalized Universal Approximation Theorem for Operators.]
\label{pythagorean}
Suppose that $X$ is a Banach space, $K_1 \subset X$, $K_2 \subset \mathbb{R}^d$ are two compact sets in $X$ and $\mathbb{R}^d$, respectively, $V$ is a compact set in $C(K_1)$. Assume that: $\mathcal{G}: V \rightarrow C(K_2)$ is a nonlinear continuous operator. Then, for any $\epsilon > 0$, there exist positive integers $m, p$, continuous vector functions $\mathbf{g}: \mathbb{R}^m \rightarrow \mathbb{R}^p$, $\mathbf{f}: \mathbb{R}^d \rightarrow \mathbb{R}^p$, and $x_1, x_2, \dots , x_m \in K_1$ such that   
\[  \Bigg\lvert \mathcal{G}(\mathbf{x})(\zeta) - \langle  \underbrace{\mathbf{g}(\mathbf{x}(x_1), \mathbf{x}(x_2), \ldots, \mathbf {x}(x_m))}_{\text{branch}}, \underbrace{\mathbf{f}(\zeta)}_{\text{trunk}}  \rangle \Bigg\rvert < \epsilon \]
holds for all $\mathbf{x} \in V$ and $\zeta \in K_2$, where $\langle \cdot, \cdot \rangle$ denotes the dot product in $\mathbb{R}^p$. For the two functions $\mathbf{g}, \mathbf{f}$ classical deep neural network models and architectures can be chosen that satisfy the universal approximation theorem of functions, such as fully-connected networks or convolutional neural networks.
\end{theorem}
The interested reader can find more information and details regarding the proof of the theorem in \cite{lu2021learning}.

\subsubsection*{Fourier neural operator}

The backbone algorithm of the Fourier neural operator (FNO) was originally introduced with the kernel integral operators in \cite{li2020neural}, while the actual model  was proposed in \cite{li2020fourier} and is based on the idea of parameterizing the integral kernel in the Fourier space. Similarly to DeepONet, FNO aims to learn a mapping between two infinite dimensional (functional) spaces. The method employs an iterative algorithm to predict a sequence of functions $v_0 \mapsto v_1 \mapsto \dots \mapsto v_T$ taking values in $\mathbb{R}^{d_v}$ formally defined as
\begin{equation}
\begin{aligned}
    v_{t+1}(x) := \sigma \Big( W v_t(x) + (\mathcal{K}(\mathbf{x}; \phi) v_t ) (x) \Big), \hspace{10pt} \forall x \in \Omega
\end{aligned}
\label{eq:iterative-FNO}
\end{equation}
where $\mathcal{K}: \mathcal{X} \times \Theta_{\mathcal{K}} \rightarrow \mathcal{H}(\mathcal{Y} (\Omega; \mathbb{R}^{d_v}), \mathcal{Y}(\Omega;\mathbb{R}^{d_v}))$ maps to bounded linear operators on $\mathcal{Y}(\Omega; \mathbb{R}^{d_v})$ and is parameterized by $\phi \in \Theta_{\mathcal{K}}, W: \mathbb{R}^{d_v} \rightarrow \mathbb{R}^{d_v}$ is a linear transformation and $\sigma: \mathbb{R} \rightarrow \mathbb{R}$ is an activation function to introduce non-linearity. The kernel integral operator $\mathcal{K}(\mathbf{x};\phi)$ is defined as
\begin{equation}
\begin{aligned}
    \big( \mathcal{K}(\mathbf{x};\phi)v_t\big)(x) := \int_{\Omega} \kappa \big(x,y,\mathbf{x}(x), \mathbf{x}(y);\phi \big)v_t(y)dy, \hspace{10pt} \forall x \in \Omega
\end{aligned}
\label{eq:kernel-integral}
\end{equation}
where $\kappa_{\phi}: \mathbb{R}^{2(d+d_x)} \rightarrow \mathbb{R}^{d_v \times d_v}$ is approximated by a neural network parameterized by $\phi \in \Theta_{\mathcal{K}}$. In FNO, the kernel integral operator in Eq.~\ref{eq:kernel-integral} is replaced with a convolution operator defined in Fourier space. The dependence on the input function $\mathbf{x}$ is removed by imposing $\kappa_{\phi}(x,y) = \kappa_{\phi}(x-y)$ and thus the operator in Eq.~\ref{eq:kernel-integral} results in
\begin{equation}
\begin{aligned}
    \big( \mathcal{K}(\mathbf{x};\phi)v_t\big)(x) = \mathcal{F}^{-1} \Big(\mathcal{F}(\kappa_{\phi}) \cdot \mathcal{F}(v_t)\Big) (x), \hspace{10pt} \forall x \in \Omega
\end{aligned}
\label{eq:fourier-integral}
\end{equation}
where $\mathcal{F}$, $\mathcal{F}^{-1}$ denote the forward and inverse Fourier transformation of a function $f: \Omega \rightarrow \mathbb{R}^{d_v}$ defined as
\begin{equation}
\begin{aligned}
    (\mathcal{F}f)_j(k) = \int_{\Omega} f_j(x) e^{-2i\pi \langle x,k \rangle}dx,  \hspace{15pt} (\mathcal{F}^{-1}f)_j(x) = \int_{\Omega} f_j(k) e^{2i\pi \langle x,k \rangle}dk,
\end{aligned}
\label{eq:fft-forward}
\end{equation}
where $k \in \Omega$ represents the frequency modes and $j=1, \dots, d_v$ with $i=\sqrt{-1}$ the imaginary unit. For implementation purposes a finite-dimensional parameterization is chosen by truncating the Fourier expansion with a maximal number of modes $k_{\text{max}} = | Z_{k_{\text{max}}} | = | \{ k \in \mathbb{Z}^d : |k_j| \le k_{\text{max},j}, \ \text{for} \ j=1, \dots, d\} |$. The low frequency modes are chosen by defining an upper-bound on the $\ell_1$-norm of $k \in \mathbb{Z}^d$. 

The complete FNO algorithm is employed as follows. An input $\mathbf{x} \in \mathcal{X}$ is first lifted to a higher dimensional representation $v_0(x) = P(\mathbf{x}(x))$ parameterized by a shallow FNN. Subsequently, a number of iterations of updates are applied $v_t \mapsto v_{t+1}$ through a series of Fourier layers. At each Fourier layer, and given that $\Omega$ is discretized with $m \in \mathbb{N}$ points we have that $v_t \in \mathbb{R}^{m \times d_v}$ and $\mathcal{F}(v_t) \in \mathbb{C}^{m \times d_v}$ which results to $\mathcal{F}(v_t) \in \mathbb{C}^{k_{\text{max}} \times d_v}$ after the truncation of the higher order modes. In practice, it has been shown that $k_{\text{max},j}=12$ perform satisfactorily for most applications. Next, the output is multiplied to a weight tensor $R \in \mathbb{C}^{k_{\text{max}} \times d_v \times d_v}$. For a uniform discretization, $\mathcal{F}$ is replaced with a Fast Fourier Transform (FFT) which greatly reduces algorithmic complexity from $\mathcal{O} (m^2)$ to $\mathcal{O} (m \log m)$. After the inverse Fourier transform the output is added to another weight matrix which is multiplied with the input i.e., $W v_t(x)$, and finally the result is passed through a non-linear activation function $\sigma (\cdot)$. After a series of $T$ Fourier layers, the PDE output $\mathbf{y}(x)=Q(v_T(x))$ is computed via the transformation of $v_T$ with $Q: \mathbb{R}^{d_v} \rightarrow \mathbb{R}^{d_y}$.

In the original work, two main FNO models are proposed: the \textbf{FNO-2D} and \textbf{FNO-3D}. In FNO-3D, 3-D convolutions are performed (in space and time) and the model maps 3D functions representing the initial time steps to 3D functions representing the full trajectory. It has been shown that FNO-3D is more expressive and leads to better accuracy for sufficient data. However, it is fixed to the training interval, so once trained, it can only predict the solution in this range but for any time-discretization. On the other hand, FNO-2D, performs 2-D convolutions together with a recurrent architecture to propagate in time. While the advantage of this approach is that the model can predict the solution for any number of time steps (and for fixed time interval $\Delta t$), it has been shown that it is less expressive and more challenging to train. For more information, the interested reader is referred to \cite{li2020fourier}.

\section*{Data generation}
\label{sec:data-generation}

\subsection*{Brittle fracture mechanics}

In this application, we consider a continuum fracture modeling method (the second-order phase field model), to approximate the growth of fracture on a unit square plate, which is fixed on the bottom and the left edge, subjected to displacement controlled shear loading conditions on the top edge [\cite{goswami2021phase}]. We specifically aim to approximate the mapping $\mathcal{G}: H(\boldsymbol{x},t=0; l_c, y_c) \mapsto \phi(\boldsymbol x,t)$. We consider the material parameters as: $\lambda = $ 121.15 kN/mm$^{2}$, $\mu = $ 80.77 kN/mm$^{2}$ and $G_c = 2.7 \times 10^{-3}$ kN/mm, where $\lambda$ and $\mu$ are Lam\'e's constants. The computation is performed by applying constant displacement increments of $\Delta u$ = $1\times 10^{-4}$ mm to effectively capture the crack propagation. For all simulations, $l_0$ is considered to be $0.0125$ mm.

\begin{figure}[ht!]
\begin{center}
\includegraphics[width=1\textwidth]{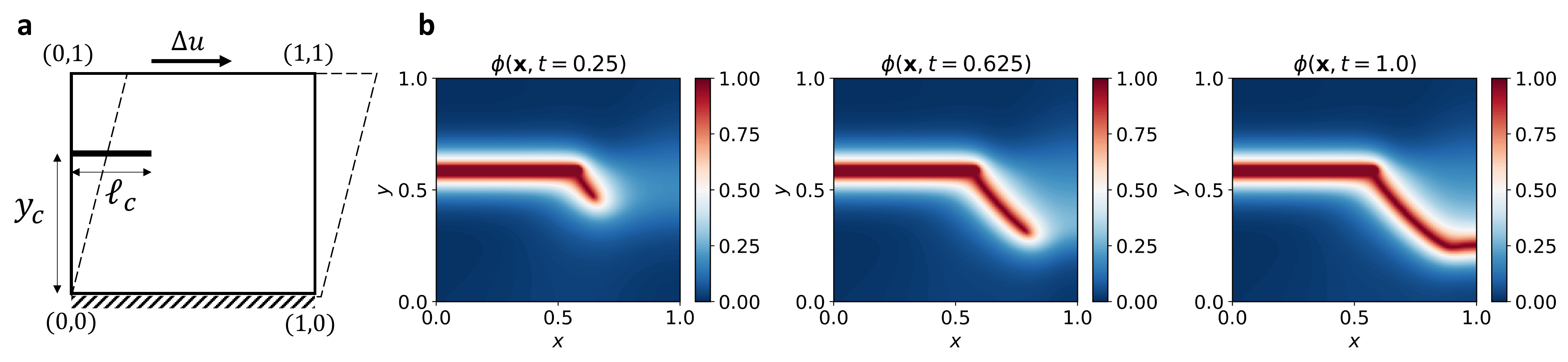}
\caption{(a) Schematic of the simulation box considered in the generating labeled dataset for brittle fracture under shear loading, depicting the two random parameters, namely the length of the crack ($l_c$) and the height of the crack ($y_c$) and (b) resulting phase field $\phi(\boldsymbol x)$ from the solution of the PDE model, showing the evolution of the crack through three-time steps $t=\{0.25,0.625,1\}$.}
\label{fig:fracture-SI}
\end{center}
\captionsetup{justification=centering}
\end{figure}

Initial cracks are modeled by using the local strain-history function, $H(\boldsymbol{x},t)$. The initial strain-history function, $H(\boldsymbol{x},t=0)$ is defined as a function of the closest distance of any point, $\boldsymbol{x}$, on the domain to the line, $l$, which represents the discrete crack [\cite{goswami2021phase}]. In particular, it is set as:
\begin{equation}\label{eq:initial_history_field}
    H(\boldsymbol{x},t=0; l_c, y_c) = \left\{ {\begin{array}{l l}
  {\frac{BG_c}{2l_0}(1 - \frac{2d(\boldsymbol{x},l)}{l_0})}&{d(\boldsymbol{x},l) \leqslant \frac{l_0}{2}} \\ 
  0&{d(\boldsymbol{x},l) > \frac{l_0}{2}}
\end{array}} \right.,
\end{equation}
where $B$ is a scalar parameter that controls the magnitude of the scalar history field and for this experiment is considered as $B=10^3$ based on domain knowledge. The function $d(\boldsymbol x, l)$ computes the distance between the middle horizontal line (defined by the two parameters $l_c, y_c$) of the crack and sets the appropriate value for the initial strain functional.  The simulation takes place in a rectangular domain $\Omega=[0,1]\times [0,1]$, discretized with $m_s \times m_s = 162 \times 162$ mesh points. The quasi-static problem is solved and in total $m_t = 8$ snapshots of the phase field $\phi(\boldsymbol x)$ are considered. Thus the dimensionality of input and output realizations is $D_{x}=26,244$ and $D_{y}=209,952$ respectively. In total, we generate $N=261$ data and split to $N_{\text{train}} = 230, N_{\text{test}}=31$ for testing and training respectively. Figure \ref{fig:fracture-SI} depicts the simulation box with the associated varying parameters as well as a representative realization of the model with the propagation of an initial crack through the phase field quantity in three points in time. The training datasets are generated using the code developed in \cite{goswami2020adaptive}, which is available on  \url{https://github.com/somdattagoswami/IGAPack-PhaseField}.

\subsection*{Rayleigh-Bénard fluid flow convection}

\begin{figure}[ht!]
\begin{center}
\includegraphics[width=0.9\textwidth]{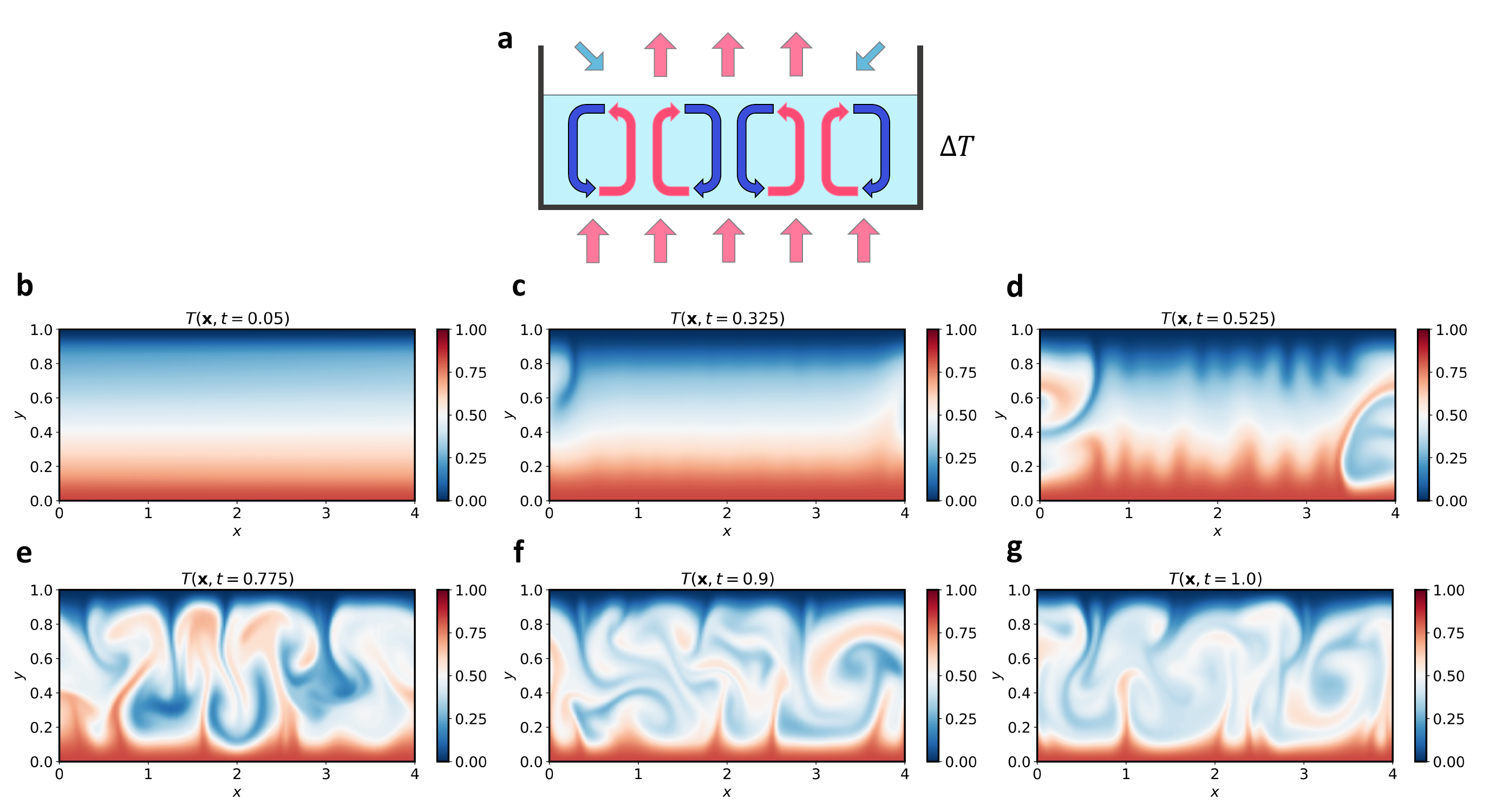}
\caption{(a) Schematic of the Rayleigh-Bénard convective flow in a thin fluid layer due to temperature gradient $\Delta T$ with the creation of convective cells at the top and (b)-(g) the evolution of the temperature field $T(\boldsymbol x, t)$ for a random realization of the initial temperature field for six time steps $t=\{0.05, 0.325, 0.525, 0.775, 0.9, 1.0\}$ based on the numerical solution of the PDE.}
\label{fig:benard-SI}
\end{center}
\captionsetup{justification=centering}
\end{figure}

In this problem, we aim to approximate the operator $\mathcal{G}: T(\boldsymbol x, t=0) \mapsto T(\boldsymbol x, t) $, which maps the initial temperature field to its entire time evolution. The simulation takes place in a spherical domain $\Omega=[0,4]\times [0,1]$, discretized with $m_s \times m_s = 128 \times 128$ mesh points. For each realization, the PDE is solved in the time interval $t=[0,1]$ for $\delta t=10^{-2}$ and $m_t=40$ times steps are considered from the 100. The dimensionless Rayleigh number is set equal to $2\cdot 10^{6}$, while the Prandtl number is set equal to $1$.
Thus the dimensionality of input and output realizations are $D_{x}=16,384$ and $D_{y}=655,360$ respectively. In total, we generate $N=800$ data and split to $N_{\text{train}} = 720, N_{\text{test}}=80$ for testing and training respectively. In Figure \ref{fig:benard-SI}, a schematic of the convective flow and a random realization of the evolution of the temperature field $T(\boldsymbol x, t)$ are shown. Datasets were generated using the \textit{Dedalus  Project} that can be found in \url{https://github.com/DedalusProject/dedalus}.
 
\subsection*{Shallow-water equations}

In this problem, we aim to approximate the operator between the random Gaussian perturbation $h'$ to the time-evolved velocity component $u$, i.e., $\mathcal{G}:h'(\lambda, \phi, t=0) \mapsto u(\phi, \lambda, t)$. The constants are defined as: $\Xi = 7.292 \times 10^{-5} \ \text{s}^{-1}$ is the Earth's angular velocity, $g=9.80616 \ \text{m} \text{s}^{-1}$ the gravitational acceleration, $\nu=1.0 \times 10^5 \ \text{m}^2 \text{s}^{-1}$ the diffusion coefficient, $u_{\text{max}}=80 \ \text{m} \text{s}^{-1}$,  $\phi_0=\pi/7$, $\phi_1=\pi/2-\phi_0$, thus the mid-point of the jet where the maximum velocity is applied is at $\phi=\pi/4$. The initial velocity $u$ is defined, so that it is zero outside the zone of interest with no discontinuities in the northern and southern poles.  The parameters  of the Gaussian perturbation which is added to the height field are set as: $\phi_2= \pi/4$, $\hat{h}=120 \ \text{m}$, while $\alpha, \beta$ are random parameters. In this expression, the Gaussian functions are multiplied with a cosine so that the forced perturbation is zero at the two poles. 

While the initial condition of the velocity field is given analytically (see Main Text), the height field is obtained by numerically integrating the balance equation
\begin{equation}
\label{eq:height-SI}
    g h(\phi) = g h_0 - \int^{\phi} \alpha u(\phi') \bigg[f + \frac{\text{tan}(\phi')}{\alpha} u(\phi') \bigg] d \phi' ,
\end{equation}
where $\alpha=6.37122 \times 10^6 \ \text{m}$ is the radius of the Earth and $h_0$ is set so that mean layer depth around the sphere is equal to $10 \ \text{km}$. The above integral can be calculated using a numerical scheme such a Gaussian quadrature. The Gaussian perturbation $h'(\lambda, \phi, t=0)$, is added to the initial height field computed by the expression above to form the final initial condition $h(\lambda, \phi, t=0)$.

\begin{figure}[ht!]
\begin{center}
\includegraphics[width=0.8\textwidth]{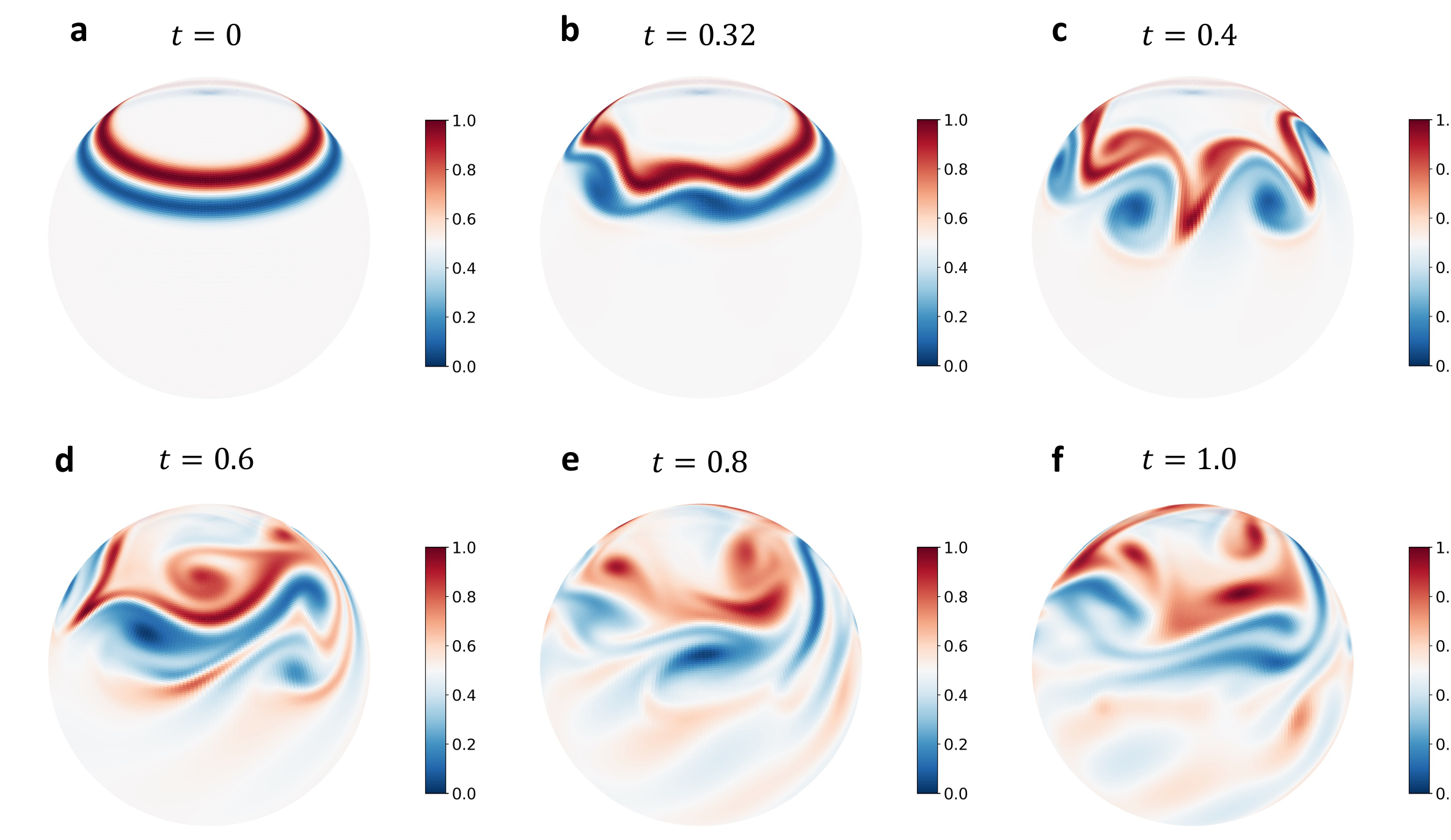}
\caption{Evolution of the velocity field $u(\lambda, \phi)$ on a sphere (Earth) as a solution of the spherical shallow-water equations, for a random realization of the initial perturbation to the height field, i.e., $\alpha=0.38, \beta=0.20$. The velocity field is shown for six time steps $t=\{0,0.32,0.4,0.6,0.8,1.0\}$.}
\label{fig:shallow-SI}
\end{center}
\captionsetup{justification=centering}
\end{figure}

The simulation takes place in a spherical domain $\Omega=[-\pi, \pi] \times [-\pi, \pi]$, discretized with $m_s \times m_s = 256 \times 256$ mesh points in the longitudinal and latitudinal direction respectively. The PDE is solved in the time interval $t=[0,360h]$ for $\delta t=1.\bar{6} \cdot 10^{-1} h$ and in total $m_t=72$ times steps (equispaced) are considered. For the presentation of results, the time range is mapped to the dimensionless range $t=[0,1]$. Thus the dimensionality of input and output realizations is $D_{x}=65,536$ and $D_{y}=4,587,520$ respectively. The significantly high dimensionality of outputs makes this problem particularly challenging. In total, we generate $N=300$ data and split to $N_{\text{train}} = 260, N_{\text{test}}=40$ for testing and training respectively. The evolution of the velocity field $u$ for a random realization of the initial height field is shown in Figure \ref{fig:shallow-SI} for six points in time. Datasets were generated using the \textit{Dedalus  Project} that can be found in \url{https://github.com/DedalusProject/dedalus}.

\section*{Network architecture details}
\label{sec:network-architecture} 

The proposed approach employs autoencoders to reduce the dimensionality of input and output PDE data and feed the DeepONet model. Although the framework is general enough and any suitable autoencoder model can be used, including convolutional autoencoders (CAE) and variational autoencoders (VAE), we found that simple multi-layer autoencoders (MLAE) resulted in the best L-DeepONet performance. Due to the large number of available training data (each output snapshot is considered a training image), all autoencoders result in very good reconstruction accuracy. However, not all autoencoders construct a latent space which is suitable for the training of DeepONet. The choice of the autoencoder also depends on the choice of the DeepONet architecture. For example, if a CNN is employed in the DeepONet's branch net, then it is not advised to use CAE for dimension reduction as the input functions will be convolved twice. 

\begin{table}[ht!]
\begin{center}
\caption{Architecture of multi-layer autoencoders (MLAE). Parameter $d$ represents the dimensionality of the latent space. All layers use the ReLU activation function except the last one which uses the sigmoid function.}
\begin{tabular}{ |c|c| } 
\hline
Application & MLAE  \\
\hline
Brittle material fracture & $[128, 64, d, 64, 128]$  \\ 
Rayleigh-Bénard fluid flow & $[400, 256, 169, d, 169, 256, 400]$   \\ 
Shallow water equation & $[256, 169, 121, d, 121, 169, 256]$   \\ 
\hline
\end{tabular}
\label{table:architectures-MLAE}
\end{center}
\end{table}

\begin{table}[ht!]
\begin{center}
\caption{Architecture of DeepONet. Inputs to the Conv2D layers consist of the number of output filters, kernel size, and activation function respectively. Parameter $p$ has been set equal to $5$.}
\begin{tabular}{ |c|c| } 
\hline
 Branch net & Trunk net \\
\hline
 \multirow{7}{*}{\begin{tabular}{@{}c@{}}Conv2D($32, (3,3)$, sine)\\Batch normalization\\Conv2D($16, (3,3)$, sine)\\Batch normalization\\Conv2D($16, (3,3)$, sine)\\Batch normalization\\Dense($p$)\end{tabular}}  &  \multirow{7}{*}{\begin{tabular}{@{}c@{}}Dense($100$) \\ Activation(sine)\\ Dense($100$) \\ Activation(sine)\\Dense($p$)\end{tabular}} \\ & \\ & \\ & \\ & \\ & \\ & \\
\hline
\end{tabular}
\label{table:architectures-DON}
\end{center}
\end{table}

Tables \ref{table:architectures-MLAE},\ref{table:architectures-DON}, show the architecture of the autoencoders and the neural operators. For all trained multi-layer autoencoders the depth and width are chosen based on the dimensionality of the original data. For the neural operators, a standard architecture is chosen which resulted in a good performance for all applications. Finally, for training both FNO-2D and FNO-3D the code from the original implementation was used which can be found at \url{https://github.com/zongyi-li/fourier_neural_operator}.

\section*{Supplementary results}
\label{sec:add-results} 

\subsection*{Error plots}

\begin{figure}[ht!]
\begin{center}
\includegraphics[width=0.6\textwidth]{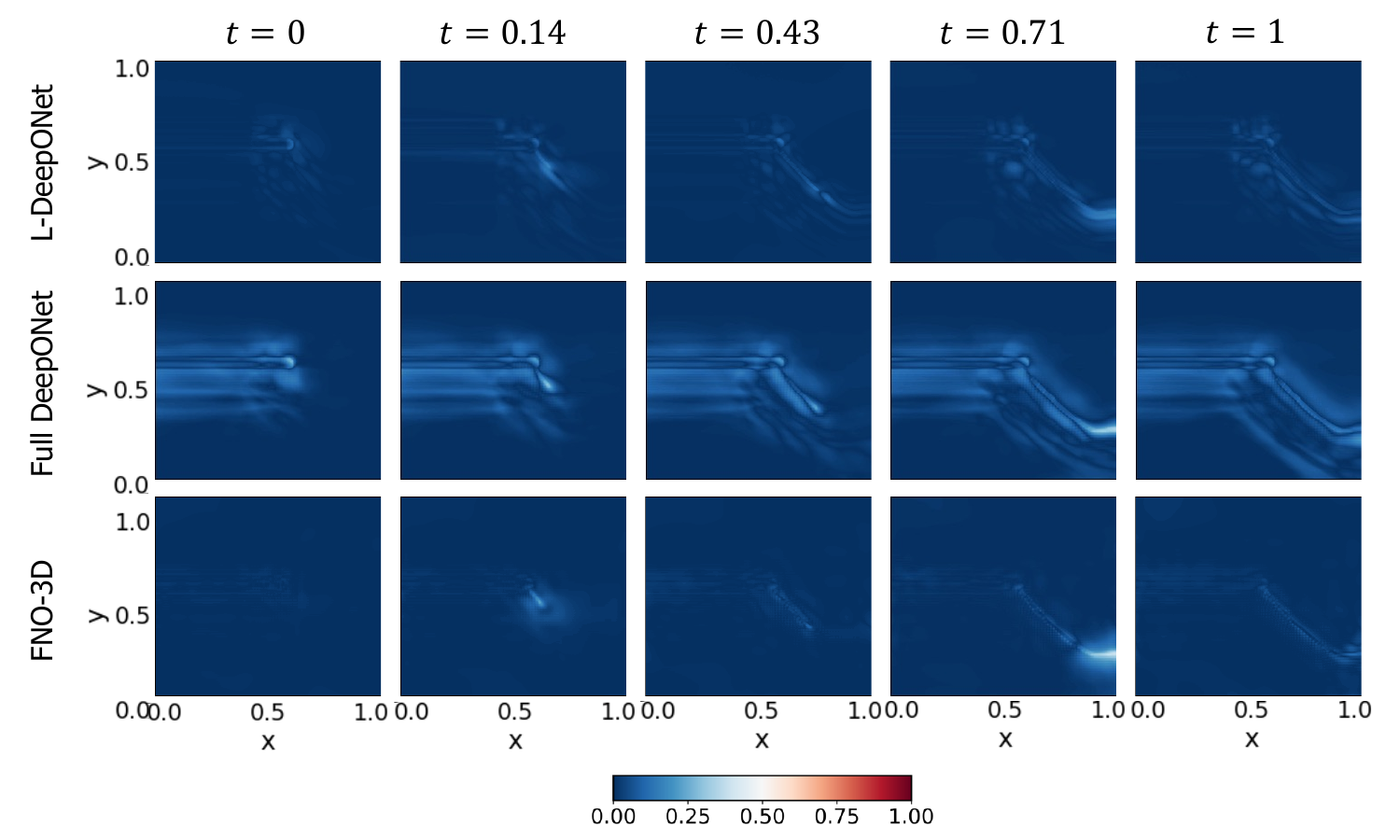}
\caption{Brittle fracture in a plate loaded in shear: absolute error plots of all the neural operators for the results of the representative sample with $y_c = 0.55$ and $l_c = 0.6$ shown in Fig. \ref{fig:fracture}. The neural operator is trained to approximate the growth of the crack for five time steps from a given initial location of the defect on a unit square domain.}
\label{fig:error-fracture-SI}
\end{center}
\captionsetup{justification=centering}
\end{figure}

\begin{figure}[ht!]
\begin{center}
\includegraphics[width=1\textwidth]{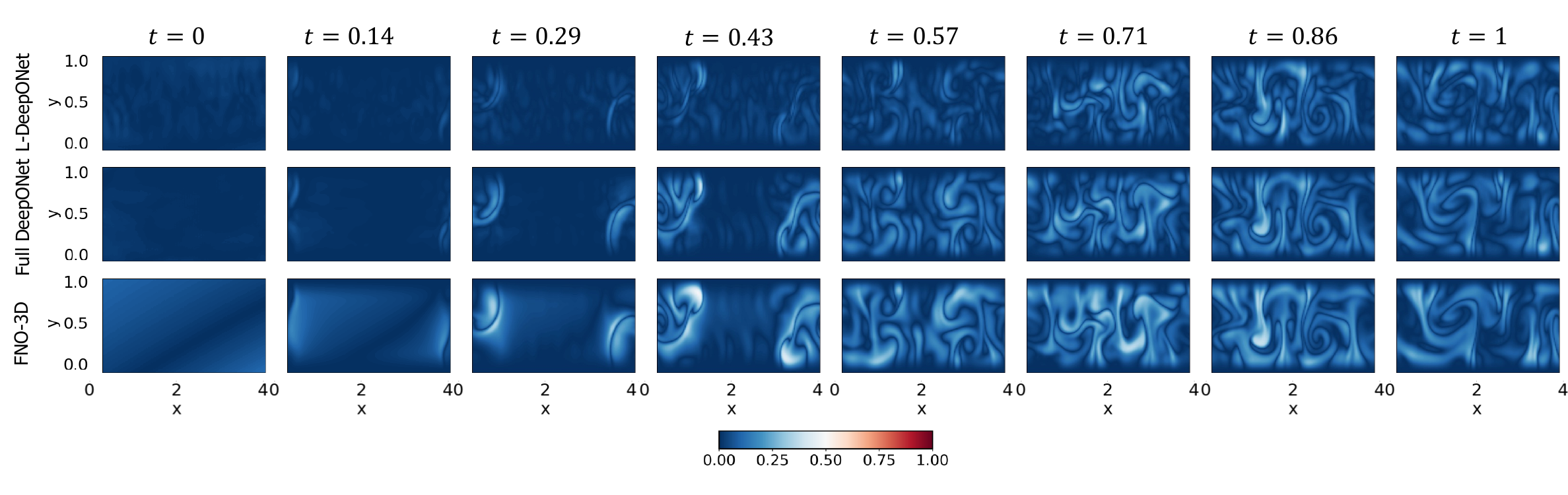}
\caption{ Rayleigh-Bénard convective flow: }
\label{fig:error-benard-SI}
\end{center}
\captionsetup{justification=centering}
\end{figure}

\begin{figure}[ht!]
\begin{center}
\includegraphics[width=0.95\textwidth]{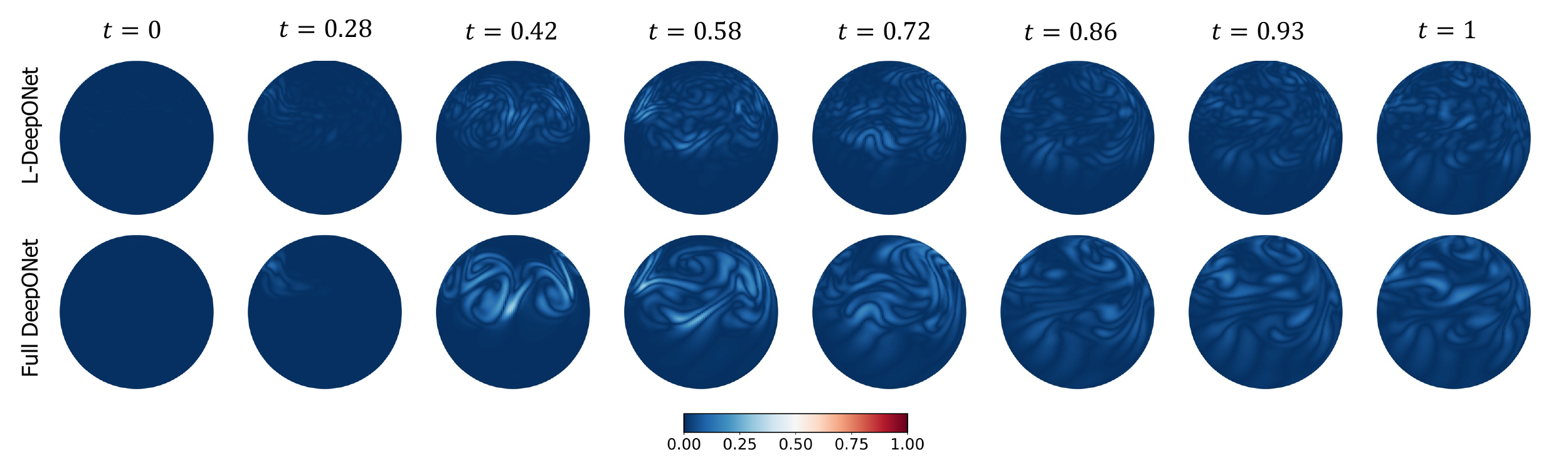}
\caption{Shallow water equations: absolute error plots of the predictions of the temperature field from a given initial temperature as obtained for all the neural operators. The predicted solution is shown in Fig. \ref{fig:benard}.}
\label{fig:error-shallow-SI}
\end{center}
\captionsetup{justification=centering}
\end{figure}

In Figures \ref{fig:error-fracture-SI},\ref{fig:error-benard-SI},\ref{fig:error-shallow-SI}, the error plots corresponding to the three applications for all studied models are presented for a single random realization. The error fields represent the point-wise absolute error between the reference response  and model prediction. As shown and discussed in the main paper, L-DeepONet results in the smallest interpolation error across diverse applications. 

\subsection*{Results using principal component analysis (PCA)}

\begin{figure}[ht!]
\begin{center}
\includegraphics[width=0.8\textwidth]{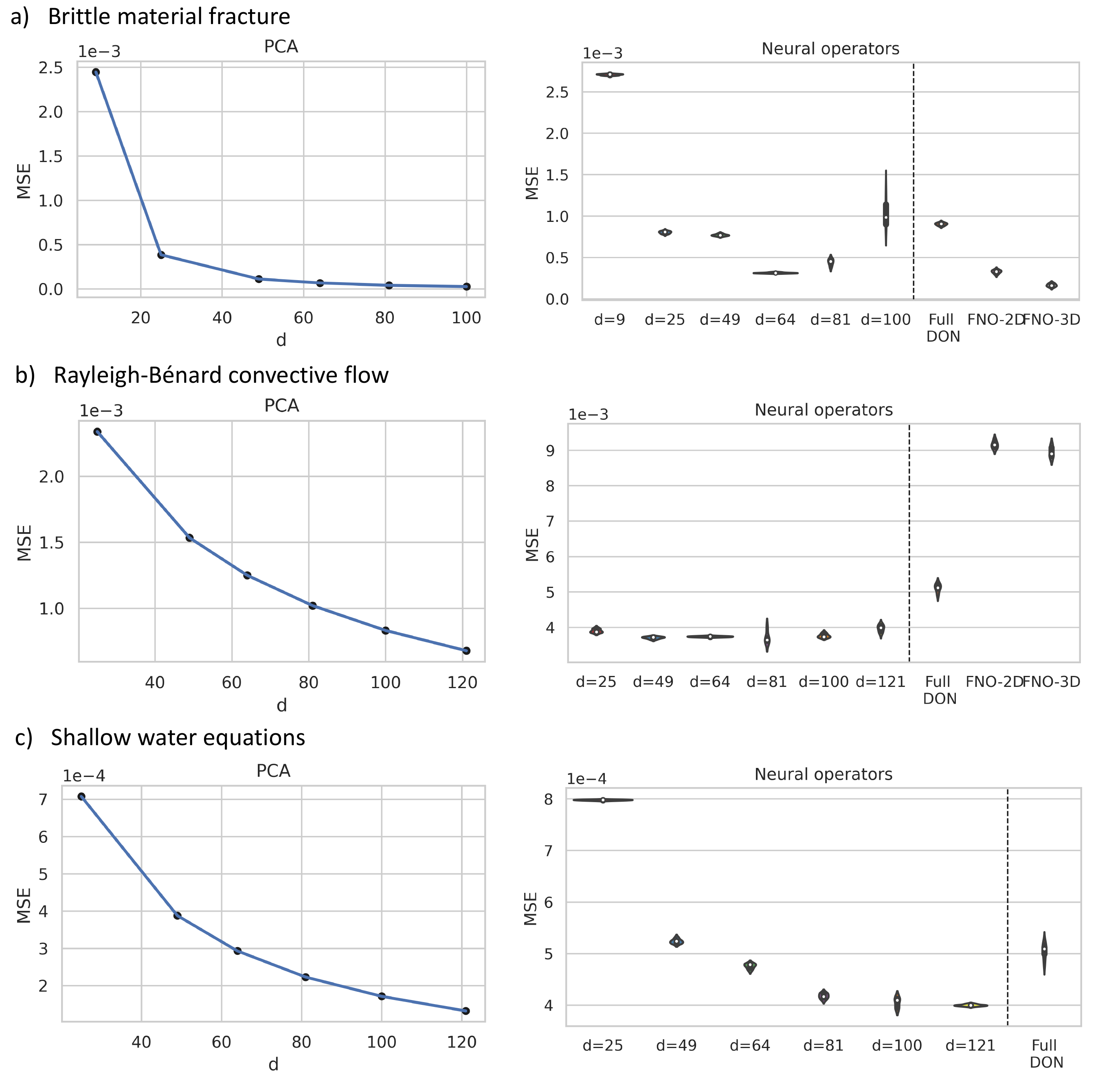}
\caption{Results for all applications of principal component analysis (PCA) (left plots) for different values of the latent dimensionality and neural operators (right plot) for all studied models. Violin plots represent 5 independent trainings of the models using different random seed numbers.}
\label{fig:PCA-L-DeepONet-results}
\end{center}
\captionsetup{justification=centering}
\end{figure}

In Figure \ref{fig:PCA-L-DeepONet-results}, we provide the results of the PCA-based L-DeepONet. In this scenario, the PCA is performed on the combined dataset of both input and output data. The left plots in Figure \ref{fig:PCA-L-DeepONet-results}, show the reconstruction MSE of the PCA method for all three PDE applications, whereas the plots on the right show the MSE of the neural operators. First, we observe that for certain problems the PCA results in low predictive accuracy for very small values of the latent dimensionality ($d=9$ in Figure \ref{fig:PCA-L-DeepONet-results} a, and $d=25$ in Figure \ref{fig:PCA-L-DeepONet-results} c). This result is also reflected in the low predictive accuracy of the neural operator model. In Figure \ref{fig:PCA-L-DeepONet-results} b, we observe that the performance of PCA and L-DeepONet when compared with the autoencoder results in the main text (Figure 2), is comparable. However, for the third and most challenging problem we found that the autoencoder (see Figure 2 in main text) outperforms the PCA-based L-DeepONet for all tested values of $d$ (see Figure \ref{fig:PCA-L-DeepONet-results} c). To summarize, we found that the autoencoder-based L-DeepONet results in a better overall performance (especially for low $d$) with an accuracy that is either comparable or better to the PCA-based L-DeepONet. However, in certain problems PCA can performance as good as the AE, with the additional advantage of being much less computationally expensive.

\end{document}